\documentclass[10pt,twocolumn,letterpaper]{article}

 \usepackage[pagenumbers]{iccv} 

\usepackage[utf8]{inputenc}
\DeclareUnicodeCharacter{0306}{}
\newcommand{\mysection}[1]{\vspace{2pt}\noindent\textbf{#1.}}
\usepackage{caption}
\usepackage{subcaption}
\usepackage{multirow}
\usepackage[table, dvipsnames]{xcolor}
\usepackage{tikz}
\usepackage{multirow}
\usepackage{float}
\usetikzlibrary{tikzmark}
\usepackage{mdframed}

\usepackage{pifont}
\newcommand{\cmark}{\ding{51}}
\newcommand{\xmark}{{\color{lightgray}\ding{55}}}
\definecolor{camelGreen}{rgb}{0.13, 0.55, 0.13}
\definecolor{babyblue}{rgb}{0.54, 0.81, 0.94}

\newcommand*\annotatedFigureBoxCustom[8]{\draw[#5,rounded corners=0.5mm, fill=#5, fill opacity=0.17, line width=0.2mm] (#1) rectangle (#2);}
\newcommand*\annotatedFigureBox[5]{\annotatedFigureBoxCustom{#1}{#2}{#3}{#4}{#5}{white}{black}{black}}

\newcommand*\annotatedRotFigureText[4]{\node[draw=none, anchor=south west, text=#2, inner sep=0, text width=#3\linewidth, rotate=90] at (#1){\footnotesize #4};}
\newenvironment {annotatedFigure}[1]{\centering\begin{tikzpicture}
\node[anchor=south west,inner sep=0] (image) at (0,0) { #1};\begin{scope}[x={(image.south east)},y={(image.north west)}]}{\end{scope}\end{tikzpicture}}

\usepackage{rotating}

\usepackage[normalem]{ulem} 
\newcolumntype{H}{>{\setbox0=\hbox\bgroup}c<{\egroup}@{}} 

\definecolor{iccvblue}{rgb}{0.21,0.49,0.74}
\usepackage[pagebackref,breaklinks,colorlinks,allcolors=iccvblue]{hyperref}

\def\paperID{} 
\def\confName{ICCV}
\def\confYear{2025}

\title{CAMELTrack: Context-Aware Multi-cue ExpLoitation for Online Multi-Object Tracking}

\author{Vladimir Somers$^1$$^2$$^3$\thanks{Equal contributions.}
\quad
Baptiste Standaert$^1$$^*$
\quad
Victor Joos$^1$$^*$
\\
Alexandre Alahi$^2$
\quad
Christophe De Vleeschouwer$^1$ \\
\\
$^1$UCLouvain \quad $^2$EPFL \quad $^3$Sportradar\\
}

\begin{document}
\maketitle
\begin{abstract}

Online multi-object tracking has been recently dominated by tracking-by-detection (TbD) methods, where recent advances rely on increasingly sophisticated heuristics for tracklet representation, feature fusion, and multi-stage matching. The key strength of TbD lies in its modular design, enabling the integration of specialized off-the-shelf models like motion predictors and re-identification. However, the extensive usage of human-crafted rules for temporal associations makes these methods inherently limited in their ability to capture the complex interplay between various tracking cues. In this work, we introduce CAMEL, a novel association module for Context-Aware Multi-Cue ExpLoitation, that learns resilient association strategies directly from data, breaking free from hand-crafted heuristics while maintaining TbD's valuable modularity. At its core, CAMEL employs two transformer-based modules and relies on a novel association-centric training scheme to effectively model the complex interactions between tracked targets and their various association cues. Unlike end-to-end detection-by-tracking approaches, our method remains lightweight and fast to train while being able to leverage external off-the-shelf models. Our proposed online tracking pipeline, CAMELTrack, achieves state-of-the-art performance on multiple tracking benchmarks. Our code is available at \url{https://github.com/TrackingLaboratory/CAMELTrack}.

\end{abstract}

\section{Introduction}
\label{sec:introduction}

\begin{figure}[t!]
\centering
\includegraphics[width=0.99\linewidth]{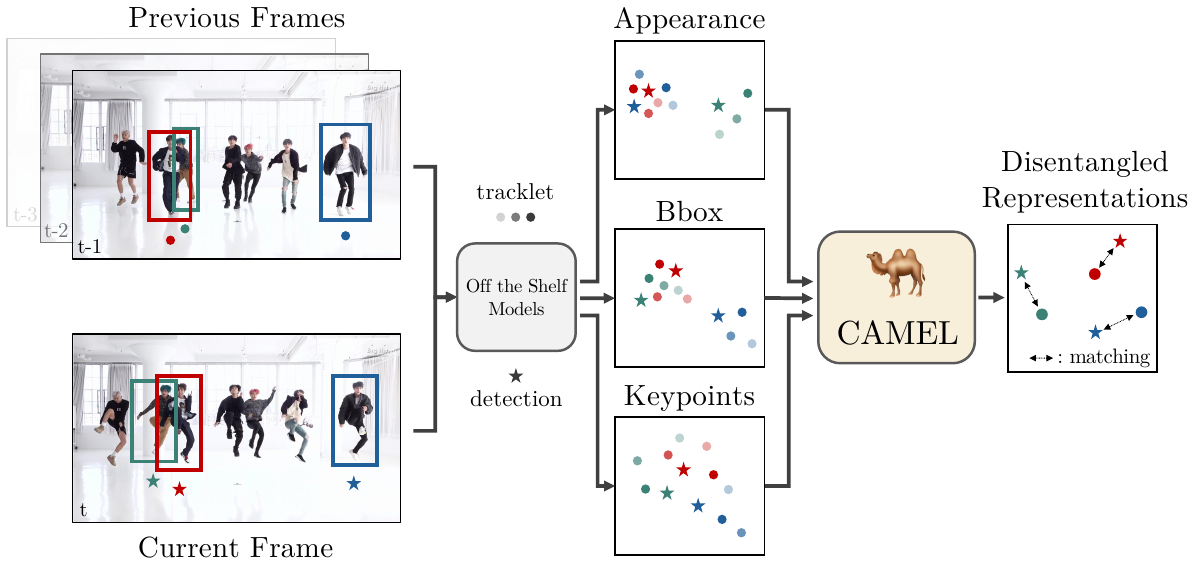}
  \caption{
  Our proposed \textit{CAMEL} association module for online tracking learns to produce disentangled tracklet and detection representations by combining various imperfect tracking cues.
  }
\label{fig:pull_figure}
\end{figure}

Multi-Object Tracking (MOT) aims to detect objects and maintain their identities across video frames, a crucial task for applications ranging from sports analytics \cite{sngamestate, soccernet22, soccernettracking, sportsmot} to autonomous driving \cite{bdd100k, adatrack}. In \textit{online MOT}, decisions must be made immediately as each frame arrives, making it challenging yet crucial for real-time processing.
The field is currently dominated by two paradigms: (i) SORT-based methods, and (ii) end-to-end (E2E) methods. 

With the emergence of powerful object detectors \cite{yolox, detr}, SORT-based \cite{sort, deepsort, bytetrack} methods, building upon the \textit{tracking-by-detection} (TbD) paradigm, have been particularly influential. Their success stems from a \textit{modular design}, where specialized components—detectors \cite{yolox}, re-identification models \cite{bpbreid, kpr}, and motion predictors \cite{diffmot, ocsort}—are independently optimized then combined through algorithmic association rules.
The \textit{association module} in SORT-based TbD pipelines, responsible for matching new detections with existing tracklets, usually encompasses three families of heuristics: (i) \textit{tracklet representation} to aggregate frame-wise detection cues over time, (ii) \textit{feature fusion} to combine multiple tracking cues into a single tracklet-detection cost matrix, and (iii) \textit{multi-stage matching} to perform sequential bipartite matching operations, each utilizing distinct cues or feature fusion strategies, and operating on specific subsets of tracklets and detections. 
\textit{Feature fusion}, the most critical component of the association module, typically relies on static combinations of motion and appearance cues \cite{hybridsort, botsort, strongsort, deepsort}.  However, as shown in \cite{ghost, deepocsort}, cue reliability fluctuates with context — particularly during occlusions, long-term associations, or when tracking visually similar targets. While some approaches attempt \textit{context-aware feature fusion} \cite{ghost, deepocsort}, their heuristic nature cannot fully capture the complex interplay between (i) the association cues and (ii) the tracked objects, suggesting the need for a more principled, data-driven approach.

To quantify the limitation of these association heuristics, we conduct an oracle-based study in~\cref{subsec:ablation} that reveals SORT-based methods fail to effectively leverage their strong association cues: when maintaining identical cues but replacing the association heuristic with an optimal oracle, HOTA improves by $15.5\%$ and $8.3\%$ on DanceTrack and SportsMOT respectively.
This demonstrates substantial room for improving association within the TbD paradigm, which remains appealing due to its ability to leverage off-the-shelf models offering strong association cues. 
To get the best out of the TbD paradigm, we propose to learn an effective context-aware association strategies directly from data, rather than designing more sophisticated heuristics.
Surprisingly, however, fully-learned association modules in online TbD remain largely unexplored. Even the transformer-based TransMOT \cite{transmot}, the most relevant prior work which made initial progress in this direction, still heavily relies on heuristics (as discussed further in \cref{sec:related_work}).

To break free from these heuristics, the majority of recent literature has shifted toward the DETR-based end-to-end (E2E) paradigm, with methods like MOTR~\cite{motr} offering a promising, data-driven alternative to TbD approaches.

Despite their elegant design with learned association, E2E methods face several limitations compared to SORT-based method, fully detailed in \cref{sec:related_work}.
A significant drawback is that E2E methods are designed to learn all subtasks (detection, reid, association) from scratch, forcing joint optimization of antagonistic objectives (a well-documented issue \cite{motrv2, motip}) while preventing the use of specialized external models. These fundamental limitations consequently require substantial training data and computational resources, typically several days of training on 8 GPUs.

Given the limitations of both E2E and SORT-based methods, we bridge the gap between the two paradigms by proposing \textbf{CAMEL}, a novel association module for \textbf{C}ontext-\textbf{A}ware \textbf{M}ulti-\textbf{C}ue \textbf{E}xp\textbf{L}oitation that replaces traditional SORT-like association heuristics with a unified trainable architecture.
CAMEL's compact and minimalist architecture consists of: (i) a set of Temporal Encoders (TE) that aggregate each tracking cue into tracklet-level representations, and (ii) a Group-Aware Feature-Fusion Encoder (GAFFE) that jointly transforms all cues into unified disentangled representations for each tracklet and detection. 
As illustrated in \cref{fig:pull_figure}, CAMEL properly discriminates matching tracklets and detections despite occlusions or similar-looking targets, by dynamically balancing multiple imperfect association cues.
This capability stems from its context-aware processing, that accounts for interactions between targets and the relative discriminativeness of each cue.
Our resulting \textit{heuristic-free} \textit{online} \textit{TbD} tracker, \textbf{CAMELTrack}, achieves state-of-the-art performance on five popular MOT benchmarks.

\noindent Overall, we summarize our contributions as follows:

\begin{itemize} 

\item We propose CAMEL which, to our knowledge, represents the first fully-learned and cue-agnostic association module for TbD pipelines, designed without bells and whistles. CAMELTrack runs at 13 FPS, which is faster than previous transformer-based trackers. 

\item We introduce an efficient Association-Centric Training, requiring under an hour on a single GPU, whereas E2E methods typically need days on multiple GPUs. 

\item We show that learned association with off-the-shelf models outperforms both E2E and SORT-based methods across five challenging benchmarks, effectively combining the strengths of both paradigms.

\end{itemize}
We release our framework and models weights to encourage further research on learned TbD association modules.

\begin{figure*}[t!]
\centering
\includegraphics[width=0.99\linewidth]{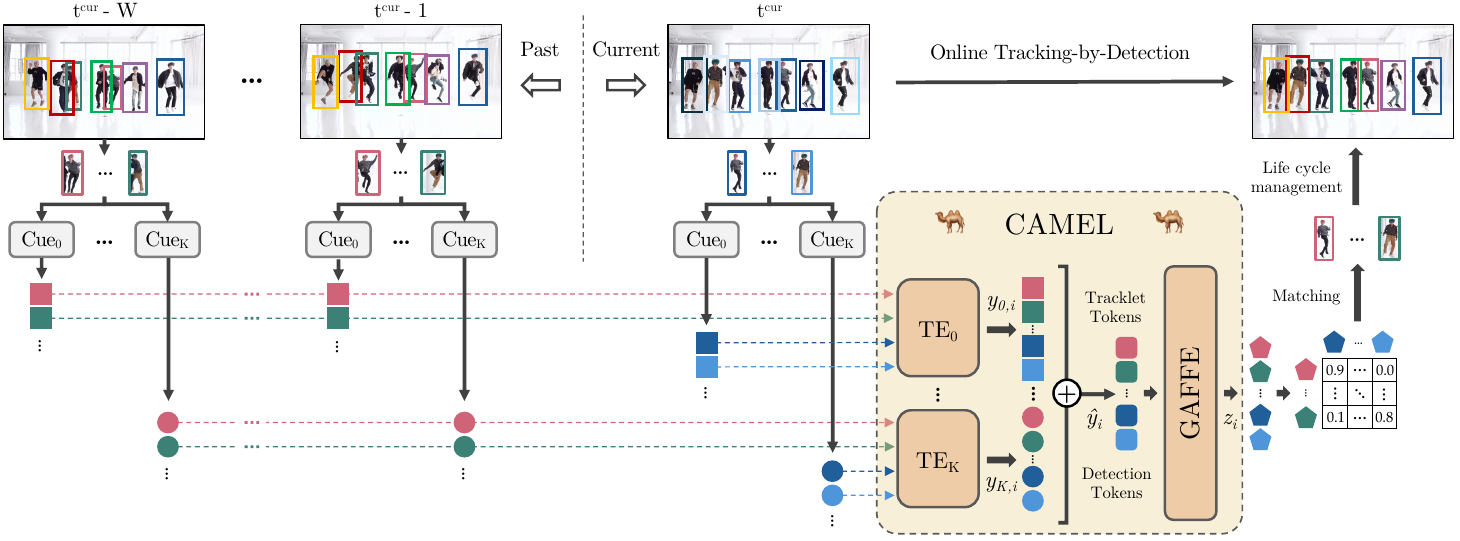}
  \caption{
  Architecture overview of CAMELTrack, our online tracking-by-detection pipeline that operates in three steps: (i) object detection, (ii) cue extraction, (iii) single-stage association using our trainable CAMEL module, and (iv) tracklet life cycle management. CAMEL processes the various imperfect cues through two stages: First, the Temporal Encoder (TE) aggregates each cue into tracklet-level representations. Second, the Group-Aware Feature Fusion Encoder (GAFFE) embeds all detection and tracklet cues into a unified discriminative embedding space. The resulting disentangled tracklets and detections representations are finally paired through bipartite matching.
  }
\label{fig:architecture}
\end{figure*}

\section{Related Work} \label{sec:related_work}

\begin{table}[t]
    \centering
    \small  
    \renewcommand{\arraystretch}{0.1} 
    \setlength{\tabcolsep}{2pt} 
    \definecolor{darkgreen}{RGB}{0,128,0}
    \definecolor{darkred}{RGB}{139,0,0}
    \centering
    \begin{tabular*}{\columnwidth}{@{\extracolsep{\fill}}c c c c c c}
        \toprule
        Paradigm & Association & Methods & HF & OM & LC \\
        \midrule
        \multirow{4}{*}{TbD} & Heuristic& SORT-based \cite{deepsort, bytetrack} & \textcolor{darkred}{\xmark} & \textcolor{darkgreen}{\cmark} & \textcolor{darkgreen}{\cmark} \\
        \cmidrule(lr){2-6}
        & Hybrid & \textit{TransMOT} \cite{transmot} & \textcolor{darkred}{\xmark} & \textcolor{darkgreen}{\cmark} & \textcolor{darkgreen}{\cmark} \\
        \cmidrule(lr){2-6}
        & \textbf{Learned} & \textbf{\textit{CAMEL} (ours)} & \textcolor{darkgreen}{\cmark} & \textcolor{darkgreen}{\cmark} & \textcolor{darkgreen}{\cmark} \\
        \cmidrule(lr){1-6}
        DbT & E2E MOT & \textit{MOTR}~\cite{motr} & \textcolor{darkgreen}{\cmark} & \textcolor{darkred}{\xmark} & \textcolor{darkred}{\xmark} \\
        \bottomrule
    \end{tabular*}
    \caption{\small Comparison of popular association paradigms and methods for Online MOT. 
    HF stands for Heuristic-Free association, OM refers to the ability to use Off-the-shelf Models, and LC denotes Low training Compute.
    }
    \label{fig:mot_taxonomy_table}
\end{table}

We review key \textit{online} MOT approaches related to our work, whose categories are summarized in \cref{fig:mot_taxonomy_table}. 

\mysection{Heuristic SORT-based Trackers}
The dominant paradigm in MOT has been tracking-by-detection (TbD), with many methods building upon SORT~\cite{sort}. These approaches focus on developing sophisticated association heuristics~\cite{deepsort, bytetrack, strongsort, botsort}, or stronger motion modeling \cite{ocsort, diffmot, motiontrack, ettrack, deepmovesort, movesort, mambatrack, mambtrack} and re-identification~\cite{ghost, smiletrack, finetrack, featuresort}. SORT-based methods primarily differ in their hand-crafted rules for association across three key components:
(i) \textit{Tracklet Representation} with mean~\cite{tracktor} or EMA \cite{TRMOT, fairmot} of detection features,
(ii) \textit{Feature Fusion} with a static \cite{ghost} or adaptive \cite{deepocsort}  weighted averaging of motion and appearance cues, or threshold-based gating \cite{strongsort, botsort},
(iii) \textit{Multi-stage Matching} with either single-stage \cite{botsort} or cascaded matching \cite{deepsort}, filtering candidates objects based on confidence scores \cite{bytetrack} or track age \cite{deepsort}.
Our method take a different direction and replaces these heuristics for data association with a unified trainable architecture, that effectively leverages all available tracking cues to produce context-aware disentangled representations to be matched in a single stage.

\mysection{Tracking-by-Detection with Learned Association}
%
%
While some previous works have explored data-driven tracking through graph networks \cite{lns, sushi} or transformers \cite{gtr}, most operate offline, with only a few pioneering works attempting to integrate learned components into online TbD pipelines~\cite{transmot, tadn, busca, strn}. 
Notably, TransMOT \cite{transmot} introduces a spatial-temporal encoder for tracklet representation and a transformer for feature fusion. However, it relies on a hand-crafted multi-stage matching pipeline, the learned components being only used in the second stage, while the first and third stages remain purely based on IoU and re-identification (ReID) heuristics.
While these works represent initial steps toward learned association, they still remain dependent on heuristics. In contrast, our approach makes a decisive break from hand-crafted rules by introducing a completely trainable association module.

\mysection{Online End-to-End} Recently, end-to-end (E2E) methods \cite{motip, motr, motrv2, trackformer, transtrack, memotr, comot, adatrack, tldmot, utm} following the Detection-by-Tracking (DbT) paradigm \cite{tracktor} have emerged as a promising, heuristic-free alternative to TbD approaches. Building upon DETR~\cite{detr} architecture, these methods jointly learn object detection and association, using track queries to re-detect past objects across frames. Despite their elegant design that learns association in a data-driven way similar to our approach, E2E methods face several limitations: (i) their detector-centric multi-frame training with short time windows struggles with long-term associations \cite{memot}, (ii) they lack TbD's modular ability to leverage specialized external models (e.g., ReID, motion, ...) \cite{motip},  (iii) the inherent conflict between detection and association objectives \cite{motrv2} in a shared model limits their overall performance and (iv) they require extensive training data and computational resources to achieve competitive performance (a few days on 8 GPUs \cite{motr}).
In contrast, our method focuses solely on learning an association strategy, requiring 
less training compute, and maintains TbD's ability to leverage off-the-shelf detection, motion, and ReID models.

\section{Methodology}
\label{sec:methodology}
In this section, we detail CAMELTrack, our proposed online tracking method. We first provide an overview of the complete tracking pipeline in \cref{subsec:CAMELTrack}. In \cref{subsec:camel}, we then detail CAMEL, our trainable Context-aware Multi-cue ExpLoitation module that learns tracklet-detection association directly from data. Finally, we describe our association-centric training scheme designed to create challenging association scenarios in \cref{subsec:train}.

\subsection{CAMELTrack Pipeline}
\label{subsec:CAMELTrack}
Our tracking pipeline, CAMELTrack, follows the online tracking-by-detection paradigm, processing each incoming frame through four sequential steps: (i) object detection, (ii) cue extraction, (iii) tracklet-detection association via our CAMEL module, and (iv) tracklet life cycle management. The following paragraphs detail one complete iteration of this process, that is illustrated in~\cref{fig:architecture}.

\mysection{Detection}
We first process the incoming video frame with timestamp $t^{\text{cur}}$ with an object detector to obtain a set of detections $\mathcal{D}$, where each detection $d^{t^{\text{cur}}}$ is represented by a bounding box and its confidence score. 

\mysection{Cue Extraction}
For each detection in $\mathcal{D}$, we extract multiple complementary cues to guide the association process, as a single cue is often insufficient for reliable tracking. 
The bounding box coordinates and confidence score constitute the first cue $c_0$, while $K$ additional cues $\{c_k\}_{k=1}^K$ are extracted by specialized off-the-shelf models. 
In this work, we employ \textit{re-identification features} and \textit{pose keypoints} as additional cues to complement the object location $c_0$. 
However, our CAMEL association module can ingest any type and number of input cues, enabling easy integration of additional domain-specific information (e.g., plate numbers for vehicle tracking). 
Each detection $d$ is thus characterized by its complete cue set, i.e. $d^{t^{\text{cur}}} = \{c^{t^{\text{cur}}}_k\}_{k=0}^K$.

\mysection{Association with CAMEL}
The association step objective is to match $M$ existing tracklets $\mathcal{T}$ with $N$ active detections $\mathcal{D}$ from the current frame.
We refer to all tracklets and detections considered for association as the set of \textit{active objects} \( \mathcal{A} = \mathcal{T} \cup \mathcal{D} \).
Each tracklet in $\mathcal{T}$ represents a unique tracked object and is composed of a sequence of detections $d^{t^\text{start}:t^\text{end}} = [d^{t^\text{start}}, \dots, d^{t^\text{end}}]$, where $t^\text{start}$ and $t^\text{end}$ indicate respectively the frame indices of the first and last detection in the tracklet. 
For each active tracklet, we maintain a feature bank storing the cues of its $W$ most recent detections allowing CAMEL to leverage a rich history of cues to counter potential noise in individual detections or id switches resulting from association errors.
CAMEL is the core contribution of our work. It takes as input all active tracklets $\{d_i^{t^\text{start}_i:t^\text{end}_i}\}$ for  $i \in \mathcal{T}$ and detections $\{d_j^{t^\text{cur}}\}$ for $j \in \mathcal{D}$, and outputs a single discriminative embedding $z$ per active object (detection and tracklet) in a shared latent space, where matched/unmatched pairs are localized close/far to each other.
%
Finally, CAMEL's disentangled representations are used to compute a cost matrix $C \in \mathbb{R}^{M \times N}$, where each entry $c_{i,j} = ||z_i - z_j||_2$ measures the Euclidean distance between the normalized embeddings $z_i$ of tracklet $i$ and $z_j$ of detection $j$.
The final assignment is then obtained through bipartite matching with the Hungarian algorithm. Any pair whose cost exceeds a specified threshold is left unmatched.
The context-aware architecture of CAMEL is detailed in \cref{subsec:camel}, and its training procedure in \cref{subsec:train}.

\mysection{Life Cycle Management}
CAMELTrack manages tracklet life cycles through a standard scheme: first, low-confidence detections are filtered out before association. Next, each matched detection extends its assigned tracklet by adding new cues to its feature bank. Unmatched high-confidence detections initialize new tracklets, while unmatched tracklets are temporarily paused and eventually terminated if they remain unmatched for an extended period.

\subsection{Our CAMEL Architecture} 

\label{subsec:camel}

In this section, we detail CAMEL, our trainable association module for Context-Aware Multi-Cue ExpLoitation, that is conceived with simplicity in mind.
As introduced in \cref{subsec:CAMELTrack}, CAMEL takes as input all cues from all active objects \( \mathcal{A} = \mathcal{T} \cup \mathcal{D} \), where \( \mathcal{T} \) includes the $M$ existing tracklets and $\mathcal{D}$ the $N$ current detections, and outputs their unified representations in a disentangled space. As a result, objects with same/distinct identities are embedded close/far to each other.
CAMEL replaces the the three key heuristics traditionally used in SORT-based association modules — \textit{tracklet representation}, \textit{feature fusion}, and \textit{multi-stage matching} — with a unified trainable architecture designed without bells and whistles.
CAMEL build upon two transformer components: the Temporal Encoder (TE) and the Group-Aware Feature Fusion Encoder (GAFFE).
First, TE performs intra-object self-attention to aggregate detection-level cues into robust tracklet-level representations, effectively replacing \textit{tracklet representation} heuristics.
Next, GAFFE by fusing multiple imperfect but complementary cues into a unified representation for each object.
Through inter-object self-attention, it replaces \textit{feature fusion} heuristics by maximizing discriminativeness between objects of different identities while enhancing similarity among objects of the same identity.
Both modules are detailed hereafter.
Finally, the need for \textit{multi-stage matching} naturally disappears as CAMEL processes all tracklets, detections, and cues at once, to perform association in a single unified stage.
In \cref{sec:comparison_architecture} 
, we detail how CAMEL's architecture fundamentally differs from existing transformer-based trackers, i.e. MOTR-like methods and TransMOT.

\mysection{Temporal Encoder (TE)} \label{subsec:te}
Each active object is processed by \( K+1 \) Temporal Encoders, each \( \text{TE}_k \) tackling a specific cue type and having a dedicated set of weights.
For a given active object \( i \in \mathcal{A} \) and cue $k$, the Temporal Encoder $\text{TE}_k$ processes the temporal sequence $c^{t^\text{start}:t^\text{end}}_{k,i} = [c^{t^\text{start}}_{k,i}, \dots, c^{t^\text{end}}_{k,i}]$ as follows.
First, each cue \(c_{k,i}^t\) in this sequence undergoes a linear transformation to produce a token \(x^t_{k,i}\). 
This critical step embeds low dimensional cues like bounding boxes into a high dimensional feature space.
Next, each token \(x^t_{k,i}\) is augmented with a sinusoidal positional encoding (\text{PE}) that encodes its relative temporal position, \ie, age, compared to the current frame timestamp $t^\text{cur}$, 
\begin{equation} 
   \tilde{x}_{k,i}^t = x_{k,i}^t + \text{PE}(t^{\text{cur}}-t).
\end{equation} 

Then, a learned \(\texttt{[CLS]}\) token is prepended to the sequence of tokens \(\tilde{x}^{t^\text{start}_i:t^\text{end}_i}_{k,i}\), and the resulting sequence is processed by a shallow multi-layer transformer encoder~\cite{bert}.

Finally, the encoded CLS token serves as the output of the TE, providing a single temporal representation $y_{k,i}$ of cue $k$ for object $i$,

\begin{equation} 
  y_{k,i} \leftarrow  
   \text{TE}_k([\texttt{[CLS]}, \tilde{x}^{t^{\text{start}}_i}_{k,i}, \dots, \tilde{x}^{t^{\text{end}}_i}_{k,i}]).
\end{equation}

Both tracklets in $\mathcal{T}$ and detections in $\mathcal{D}$ undergo temporal encoding—even if detections are sequences length of one—to ensure all cues are embedded in the same latent space for further processing by GAFFE.

\mysection{Group-Aware Feature Fusion Encoder (GAFFE)} \label{subsec:gaffe}
This module receives as input the temporally encoded tokens \( y_{k,i} \) produced by the Temporal Encoders, where each token corresponds to a different cue of each active object \( i \in \mathcal{A} \). GAFFE processes these tokens in two stages to produce a single discriminative embedding per object.

In the first stage, each cue-specific token \( y_{k,i} \) is linearly projected into a higher-dimensional space. The projected tokens are then fused through summation to form a single multi-modal token \( \hat{y}_i \) per active object,

\begin{equation}
\hat{y}_i = \sum_{k=0}^K \text{Linear}_k(y_{k,i}).
\end{equation}

In the second stage, the resulting sequence of \( N + M \) multi-modal tokens \( \hat{y}_i \) is processed through a shallow multi-layer transformer encoder~\cite{bert}, that performs group-aware inter-object self-attention,

\begin{equation}
\{z_i\} \leftarrow \text{GAFFE}(\{\hat{y}_i\}), \quad \forall i \in \mathcal{A}.
\end{equation}

These resulting embeddings \( \{z_i\} \) are the final, disentangled representations of each active object, which are then used for matching as detailed in \cref{subsec:CAMELTrack}.

\subsection{Association-Centric Training}  \label{subsec:train}


Existing end-to-end (E2E) methods employ a \textit{recursive multi-frame training scheme} \cite{motr,motip}, where the model processes a short video sequence frame-by-frame to learn detection and association jointly. In contrast, our proposed Association-Centric Training (ACT) strategy decouples association from both detection and cue extraction, and works as follows.
First, we generate an image-free training set by (i) running an off-the-shelf detector on all training sequences, (ii) assigning each detection the label of its IoU-closest ground-truth, (iii) extracting all required cues (e.g., re-identification, pose). 
Then, during training, we sample from our pre-generated set to build batches of $B$ training samples. Each training sample corresponds to one input of CAMEL and models a single association scenario with $P$ tracklet-detection pairs. A single scenario is constructed by choosing a random frame, collecting all detections from that frame along with tracklets from previous frames. We repeat this process with frames from distinct videos until $P$ pairs are sampled. This cross-video sampling to generate artificial association examples increases training diversity and empirically yields more stable training and faster convergence. We further enrich training by applying three data augmentations to generate more challenging and diverse association scenarios: (i) \textit{detection identity swapping}, (ii) \textit{detection dropout}, and (iii) \textit{cue dropout} (all detailed in \cref{sec:detailed_training}). Finally, we employ the InfoNCE loss~\cite{infonce} as a training objective to minimize/maximize distances between detection-tracklet pairs of same/different identities. 

ACT offers two key advantages over recursive training strategies. 
First, E2E methods are computationally constrained to short sequences due to their heavy image processing architectures. 
In contrast, our lightweight processing of pre-computed features enables efficient modeling over large time windows, thereby improving long-term tracking. 
Second, ACT's data augmentations generate synthetic training samples that model diverse challenging scenarios: occlusions, similar-looking targets, scene re-entries, noisy features, and detection errors. As demonstrated in~\cref{subsec:ablation}, exposure to these hard examples significantly improves performance.
%


\section{Experiments}

\subsection{Datasets and Metrics}

We evaluate CAMELTrack on five datasets. DanceTrack~\cite{dancetrack} features complex dancing scenarios, while SportsMOT~\cite{sportsmot} focuses on team sports players. Both benchmarks present complementary tracking challenges with comprehensive training/testing splits. MOT17 remains a well-established dataset, though recent works \cite{transmot, motrv2, motip, memotr, motr} highlight limitations for evaluating learned association approaches. In \cref{sec:extended_results}, we evaluate on the well-established pose tracking benchmark PoseTrack21~\cite{posetrack21} and on the challenging BEE24~\cite{bee24} MOT dataset.
Finally, we use HOTA~\cite{hota}, MOTA~\cite{mota} and IDF1~\cite{idf1} for evaluation. We focus our analysis on association-related metrics (AssA \& IDF1) as they directly evaluate our contribution's impact, independently of detection quality.

\subsection{Implementation details} \label{sec:implementation_details}
We use the YOLOX~\cite{yolox} detector provided by DiffMOT~\cite{diffmot}. 
For tracking cues, we leverage dataset-specific BPBReID~\cite{bpbreid} models for appearance, off-the-shelf RTMPose~\cite{rtmpose} for pose estimation.
Our tracking pipeline is implemented with TrackLab~\cite{tracklab}.
Our model employs 4-layer, 8-head transformer encoders for both TEs and GAFFE, for a total $42.6$M parameters. Training occurs over 10 epochs. 
A training sample has $P=32$ detection-tracklet pairs.
We first pretrain the TEs independently before jointly optimizing with GAFFE. Training CAMEL takes one hour on a single consumer-grade GPU.
The entire pipeline runs on average at 13 FPS on MOT17: 
24.4ms for YOLOX, 16.8ms for RTMPose, 16ms for BPBReID, and 18ms for CAMELTrack.
We employ a feature bank of $W=50$. Additional details are available in 
\cref{sec:extended_implementation_details}.

\subsection{Comparison to State-of-the-Art}

Our method establishes new state-of-the-art performance across most benchmarks, surpassing both end-to-end (E2E) methods~\cite{motr,motip} that traditionally dominate DanceTrack, and SORT-based approaches~\cite{hybridsort, diffmot} that excel on SportsMOT. Furthermore, CAMELTrack outperforms all existing learned methods~\cite{trackformer,motr} on MOT17, while achieving competitive performance with heuristic methods~\cite{centertrack,fairmot}. CAMELTrack also outperforms state-of-the-art by +7.6\% HOTA on PoseTrack21 and +3.7\% on BEE24.

\mysection{DanceTrack}
\begin{table}[t]
    \centering
    \renewcommand{\arraystretch}{0.9} 
    \resizebox{\columnwidth}{!}{
    \begin{tabular}{@{\hspace{0.5em}}l*{5}{@{\hspace{0.7em}}c}}\toprule
        Method          & HOTA$\uparrow$ & AssA$\uparrow$ & DetA$\uparrow$ & IDF1$\uparrow$ & MOTA$\uparrow$ \\\midrule
        \hspace{-0.5em}{\scriptsize \color{gray} End-to-End MOT} \\
        MOTR \cite{motr}  & 54.2 & 40.2 & 73.5 & 51.5 & 79.7\\
        MOTIP \cite{motip}  & 67.5 & 57.6 & 79.4 & 72.2 & 90.3\\
        MeMOTR \cite{memotr}  & 68.5 & 58.4 & 80.5 & 71.2 & 89.9\\
        \hline
        \hspace{-0.5em}{\scriptsize \color{gray} Heuristic Association} \\

        \rowcolor{babyblue!20}
        ByteTrack \cite{bytetrack} & 47.7 & 32.1 & 71.0 & 53.9 & 91.3\\
        \rowcolor{babyblue!20}
        OC-SORT \cite{ocsort} & 55.1 & 38.0 & 80.3 & 54.2 & 89.4\\
        \rowcolor{babyblue!20}
        GHOST \cite{ghost} & 56.7 & 39.8 & 81.1 & 57.7 & 89.6\\
        \rowcolor{babyblue!20}
        Deep OC-SORT \cite{deepocsort} & 61.3 & 45.2 & 82.2 & 61.5 & 92.3\\
        \rowcolor{babyblue!20}
        DiffMOT \cite{diffmot} & 62.3 & 48.8 & \textbf{82.5} & 64.0 & \textbf{92.7}\\
        \rowcolor{babyblue!20}
        Hybrid-SORT \cite{hybridsort}& 65.7 & -    &  -   & 67.4 & 91.8\\
        \hline
        \hspace{-0.5em}{\scriptsize \color{gray} Learned Association} \\
        \rowcolor{babyblue!20}
        CAMELTrack           & 66.1 & 54.0 & 81.1 & 71.1 & 91.4\\ 
        \rowcolor{babyblue!20}
        w/ keypoints & \textbf{69.3} & \textbf{58.9} & 81.8 & \textbf{74.9} & 91.4\\
        \bottomrule
    \end{tabular}
    }
    \caption{Comparison on DanceTrack~\cite{dancetrack} test set. For fair comparison, we only report methods trained exclusively on DanceTrack. Methods with {\color{babyblue} blue} background use the same YOLOX detector.} 
    \label{tab:dancetrack_sota}
\end{table}
As depicted in~\cref{tab:dancetrack_sota}, E2E methods~\cite{motr,motip,memotr} dominate this benchmark, outperforming existing SORT-based methods~\cite{hybridsort, diffmot, deepocsort, ghost, ocsort, bytetrack}. 
The poor performance of these SORT-based methods can be attributed to DanceTrack's challenging scenarios --- similar-looking dancers executing complex movements with frequent occlusions --- which yield unreliable motion and appearance cues, as demonstrated by our oracle-based study in~\cref{subsec:oracle}. 
Heuristic-based association is inherently more sensitive to such unreliable inputs: incorrect associations therefore occur, progressively degrading tracklets' representations and cascading into even more tracking errors.
While Hybrid-SORT~\cite{hybridsort}, attempts to address these issues by introducing three additional cues, it still remains limited by a static feature fusion.
In contrast, our data-driven association bridges the performance gap with E2E methods by learning to leverage each cue's discriminative power.
Similar to our approach, MeMOTR \cite{memotr} and MOTIP's \cite{motip} success can be attributed to their learned association.

Finally, previous attempts~\cite{dancetrack} at leveraging keypoints achieved only marginal gains ($+0.4\%$ HOTA), likely due to hand-crafted rules' limitations in exploiting this rich information. In contrast, our method yields significant improvements ($+3.2\%$ HOTA), surpassing E2E performance, while maintaining similar inference speed since RTMPose is fast.

\mysection{SportsMOT}
As reported in~\cref{tab:sportsmot_sota}, SORT-based~\cite{motiontrack, sportsmot, diffmot, deepeiou} methods dominate SportsMOT's leaderboard, outperforming E2E solutions~\cite{memotr,motip}. This success can be attributed to appearance and motion cues being more reliable on SportsMOT than on DanceTrack. For instance, even though players wear similar team uniforms, our ablation study in~\cref{subsec:ablation} demonstrates that appearance remains a very effective cue for sports tracking. 
The effectiveness of these distinguishing cues particularly benefits TbD methods, as their dedicated ReID models capture object appearance better than E2E track-queries.
On the other hand, we outperform SORT-based methods for similar reasons than DanceTrack. 
Our Association-Centric Training exposes the model to long-term associations, which improves handling of scene re-entries. 
Overall, CAMELTrack achieves significant improvements (+3.2\% HOTA) over prior state-of-the-art methods. However, unlike DanceTrack, keypoints degrade performance on SportsMOT, likely due to more distant viewpoints resulting in noisy pose estimation.
\begin{table}[t]
    \centering
    \renewcommand{\arraystretch}{0.9} 
    \resizebox{\columnwidth}{!}{
    \begin{tabular}{l*{5}{c}}\toprule
        Method          & HOTA$\uparrow$ & AssA$\uparrow$ & DetA$\uparrow$ & IDF1$\uparrow$ & MOTA$\uparrow$ \\\midrule
        \hspace{-0.5em}{\scriptsize \color{gray} End-to-End MOT} \\
        MeMOTR \cite{memotr} & 70.0 & 59.1 & 83.1 & 71.4 & 91.5 \\
        MOTIP            & 71.9 & 62.0 & 83.4 & 75.0 & 92.9 \\
        \hline
        \hspace{-0.5em}{\scriptsize \color{gray} Heuristic Association} \\
        \rowcolor{babyblue!20}
        ByteTrack \cite{bytetrack} & 62.1 & 50.5 & 76.5 & 69.1 & 93.4 \\
        \rowcolor{babyblue!20}
        OC-SORT \cite{ocsort} & 68.1 & 54.8 & 84.8 & 68.0 & 93.4 \\
        MotionTrack \cite{motiontrack} & 74.0 & 61.7 & 88.8 & 74.0 & 96.6 \\
        \rowcolor{babyblue!20}
        MixSORT  \cite{sportsmot} & 74.1 & 62.0 & 88.5 & 74.4 & 96.5 \\
        \rowcolor{babyblue!20}
        DiffMOT\cite{diffmot} & 76.2 & 65.1 & \textbf{89.3} & 76.1 & \textbf{97.1}	\\
        
        Deep-EIoU \cite{deepeiou} & 77.2 & 67.7 & 88.2 & 79.8 & 96.3 \\
        \hline
        \hspace{-0.5em}{\scriptsize \color{gray} Learned Association} \\
        \rowcolor{babyblue!20}
        CAMELTrack           & \textbf{80.4} & \textbf{72.8} & 88.8 & \textbf{84.8} & 96.3 \\
        \rowcolor{babyblue!20}
        w/ keypoints & 80.3 & 72.6 & 89.0 & \textbf{84.8} & 96.4 \\
        \bottomrule
    \end{tabular}
    }
    \caption{Comparison on SportsMOT~\cite{sportsmot} test set. Methods with {\color{babyblue} blue} background use the same YOLOX detector.}
    \label{tab:sportsmot_sota}
\end{table}

\begin{table}[t]
    \centering
    \renewcommand{\arraystretch}{0.9} 
    \resizebox{\columnwidth}{!}{
    \begin{tabular}{@{\hspace{0.5em}}l*{6}{@{\hspace{0.7em}}c}}\toprule
        Method          & HOTA$\uparrow$ & AssA$\uparrow$ & DetA$\uparrow$ & IDF1$\uparrow$ & MOTA$\uparrow$ & FPS$\uparrow$\\\midrule
        \hspace{-0.5em}{\scriptsize \color{gray} End-to-End MOT} \\
        MOTR \cite{motr}            & 57.8 & 55.7 & 60.3 & 68.6 & 73.4 & 7.5\\
        MeMOTR \cite{memotr}        & 58.8 & 58.4 & 59.6 & 71.5 & 72.8 & - \\
        MOTIP \cite{motip}          & 59.2 & 56.9 & 62.0 & 71.2 & 75.5 & - \\
        MOTRv2 \cite{motrv2}        & 62.0 & 60.6 & \textbf{63.8} & 75.0 & 78.6 & 6.9\\
        \hline
        \hspace{-0.5em}{\scriptsize \color{gray} Heuristic Association} \\
        FairMOT \cite{fairmot}          & 59.3 & - & - & 72.3 & 73.7 & 26\\
        \rowcolor{babyblue!20}
        OC-SORT \cite{ocsort}       & 61.7 & - &  -   & 76.2 & 76.0 & 28\\  
        \rowcolor{babyblue!20}
        ByteTrack \cite{bytetrack}       & \textbf{62.8} & - &  -   & \textbf{77.1} & \textbf{78.7} & 30 \\  
        \rowcolor{babyblue!20}
        GHOST \cite{ghost}          & \textbf{62.8} &  -   &  -   & \textbf{77.1} & \textbf{78.7} & 6 \\
        \hline
        \hspace{-0.5em}{\scriptsize \color{gray} Hybrid Association} \\
        TADN \cite{tadn}          & - & - & - & 60.8 & 69.0 & 10\\
        TransMOT \cite{transmot}          & - & - & - & 76.3 & 76.4 & 10\\
        \hline
        \hspace{-0.5em}{\scriptsize \color{gray} Learned Association} \\
        \rowcolor{babyblue!20}
        CAMELTrack          & 62.4 & \textbf{61.4} & 63.6 & 76.5 & 78.5 & 13\\
        \bottomrule
    \end{tabular}
    }
    \caption{Comparison on MOT17~\cite{mot17} test set on the private detection setting. Only \textbf{fully online methods} are reported for fairness. Methods with {\color{babyblue} blue} background use the same YOLOX detector.
    }
    \label{tab:mot17_sota}
\end{table}

\mysection{MOT17} Test set results are reported in \cref{tab:mot17_sota}. 
End-to-end (E2E) approaches, which jointly learn detection and association, require substantial training data~\cite{motip}. Most of these methods leverage the CrowdHuman~\cite{crowdhuman} dataset for joint training to overcome this limitation. Despite not using additional training data, CAMEL still outperforms these E2E approaches.
As detailed in \cref{sec:related_work}, TransMOT~\cite{transmot} and TADN~\cite{tadn} represent initial attempts to integrate learned components into TbD pipelines. Our approach outperforms both methods. We attribute this to our fundamentally different architecture and training on longer sequences compared to their limited 5-frame training windows.
Additionally, CAMEL achieves faster inference by using only $M+N$ object-centric tokens, avoiding the quadratic complexity of the $M\times N$ edge-centric tokens in their graph-inspired architectures (details in \cref{sec:comparison_architecture}).
SORT-based methods, have long dominated the MOTChallenge benchmark. As discussed in \cref{sec:mot17_discussion}, the structure of the dataset inherently favors such handcrafted methods, as they require only a small training set to optimize their hyper-parameters. Despite MOT17's inherent bias towards such methods, our learned CAMELTrack achieves competitive performance.

\subsection{Ablation Studies} \label{subsec:ablation}
We conduct extensive experiments in \cref{tab:main_ablation_study} on SportsMOT and DanceTrack validation sets to analyze CAMEL's design. Our study evaluates three key aspects: (i) Temporal Encoders versus standard tracklet representations heuristics (Exp. 1-5), (ii) our Group-Aware Feature Fusion Encoder (Exp. 6-8), and (iii) our complete architecture (Exp. 9-10). Additionally, we design oracle experiments (Exp. 11-12) to establish performance upper bounds.

\begin{table}[t]
  \centering
  \setlength{\tabcolsep}{5pt}  
\resizebox{\columnwidth}{!}{
\begin{tabular}{c c c c c c c c c c}
\toprule
\multirow{2.4}{*}{Exp} & \multicolumn{3}{c}{Features} & \multirow{2.4}{*}{GAFFE} & \multirow{2.4}{*}{DA} & \multicolumn{2}{c}{DanceTrack} & \multicolumn{2}{c}{SportsMOT} \\
\cmidrule(lr){2-4} \cmidrule(lr){7-8} \cmidrule(lr){9-10}
 & App & Bb & Kp & & & HOTA$\uparrow$ & IDF1$\uparrow$ & HOTA$\uparrow$ & IDF1$\uparrow$ \\
\midrule
1 & EMA    & -      & -      & \xmark & \xmark & 49.9\tikzmark{1_dt_hota} & 48.0\tikzmark{1_dt_idf1} & 76.0\tikzmark{1_sm_hota} & 80.8\tikzmark{1_sm_idf1} \\
2 & TE     & -      & -      & \xmark & \cmark & 52.4\tikzmark{2_dt_hota} & 52.6\tikzmark{2_dt_idf1} & 79.2\tikzmark{2_sm_hota} & 84.7\tikzmark{2_sm_idf1} \\
3 & -      & KF     & -      & \xmark & \xmark & 54.3\tikzmark{3_dt_hota} & 56.3\tikzmark{3_dt_idf1} & 72.1\tikzmark{3_sm_hota} & 72.6\tikzmark{3_sm_idf1} \\

4 & -      & TE     & -      & \xmark & \cmark & 54.7\tikzmark{4_dt_hota} & 57.4\tikzmark{4_dt_idf1} & 71.1\tikzmark{4_sm_hota} & 75.4\tikzmark{4_sm_idf1} \\ 
5 & -      & -      & TE     & \xmark & \cmark & 56.0 & 59.5 & 71.3 & 75.7 \\
\cmidrule(lr){1-10}

6 & EMA    & KF     & -      & \xmark & \xmark & 54.3\tikzmark{6_dt_hota} & 57.2\tikzmark{6_dt_idf1} & 75.8\tikzmark{6_sm_hota} & 80.7\tikzmark{6_sm_idf1} \\

7 & EMA    & KF     & -      & \cmark & \cmark & 56.5\tikzmark{7_dt_hota} & 58.2\tikzmark{7_dt_idf1} & 79.4\tikzmark{7_sm_hota} & 85.1\tikzmark{7_sm_idf1} \\

8 & TE     & TE     & -      & \cmark & \cmark & 62.4 & 65.7 & 81.8 & 87.9 \\
\cmidrule(lr){1-10}

9 & TE     & TE     & TE     & \cmark & \xmark & 61.0\tikzmark{9_dt_hota} & 64.9\tikzmark{9_dt_idf1} & 78.2\tikzmark{9_sm_hota} & 83.7\tikzmark{9_sm_idf1} \\
10 & TE     & TE     & TE     & \cmark & \cmark & 65.1\tikzmark{10_dt_hota} & 70.5\tikzmark{10_dt_idf1} & 81.9\tikzmark{10_sm_hota} & 88.5\tikzmark{10_sm_idf1} \\
\midrule[\heavyrulewidth]

11 & \multicolumn{5}{c}{\color[RGB]{0,80,0}Oracle Feature Fusion (KF \& EMA)} & {\color[RGB]{0,80,0}69.8} & {\color[RGB]{0,80,0}74.7} & {\color[RGB]{0,80,0}84.1} & {\color[RGB]{0,80,0}91.0} \\
12 & \multicolumn{5}{c}{\color[RGB]{0,80,0}Oracle Association} & {\color[RGB]{0,80,0}86.1} & {\color[RGB]{0,80,0}98.2} & {\color[RGB]{0,80,0}90.8} & {\color[RGB]{0,80,0}99.4} \\
\bottomrule
\end{tabular}
\begin{tikzpicture}[overlay,remember picture]
\draw[-latex,ForestGreen] ([yshift=0.3em,xshift=0.1em]{pic cs:1_dt_hota})
to [out=-45,in=45]
node[midway, right, xshift=-0.3em, yshift=+0.3em] {\tiny +2.5}
([yshift=0.4em,xshift=0.0em]{pic cs:2_dt_hota});
\draw[-latex,ForestGreen] ([yshift=0.3em,xshift=0.1em]{pic cs:1_dt_idf1})
to [out=-45,in=45]
node[midway, right, xshift=-0.3em, yshift=+0.3em] {\tiny +4.6}
([yshift=0.4em,xshift=0.em]{pic cs:2_dt_idf1});
\draw[-latex,ForestGreen] ([yshift=0.3em,xshift=0.1em]{pic cs:1_sm_hota})
to [out=-45,in=45]
node[midway, right, xshift=-0.3em, yshift=+0.3em] {\tiny +3.2}
([yshift=0.4em,xshift=0.em]{pic cs:2_sm_hota});
\draw[-latex,ForestGreen] ([yshift=0.3em,xshift=0.1em]{pic cs:1_sm_idf1})
to [out=-45,in=45]
node[midway, right, xshift=-0.3em, yshift=+0.3em] {\tiny +3.9}
([yshift=0.4em,xshift=0.em]{pic cs:2_sm_idf1});
\draw[-latex,ForestGreen] ([yshift=0.3em,xshift=0.1em]{pic cs:3_dt_hota})
to [out=-45,in=45]
node[midway, right, xshift=-0.3em, yshift=+0.3em] {\tiny +0.4}
([yshift=0.4em,xshift=0.em]{pic cs:4_dt_hota});
\draw[-latex,ForestGreen] ([yshift=0.3em,xshift=0.1em]{pic cs:3_dt_idf1})
to [out=-45,in=45]
node[midway, right, xshift=-0.3em, yshift=+0.3em] {\tiny +1.1}
([yshift=0.4em,xshift=0.em]{pic cs:4_dt_idf1});
\draw[-latex,OrangeRed] ([yshift=0.3em,xshift=0.1em]{pic cs:3_sm_hota})
to [out=-45,in=45]
node[midway, right, xshift=-0.3em, yshift=+0.3em] {\tiny -1.0}
([yshift=0.4em,xshift=0.em]{pic cs:4_sm_hota});
\draw[-latex,ForestGreen] ([yshift=0.3em,xshift=0.1em]{pic cs:3_sm_idf1})
to [out=-45,in=45]
node[midway, right, xshift=-0.3em, yshift=+0.3em] {\tiny +2.8}
([yshift=0.4em,xshift=0.em]{pic cs:4_sm_idf1});
\draw[-latex,ForestGreen] ([yshift=0.3em,xshift=0.1em]{pic cs:6_dt_hota})
to [out=-45,in=45]
node[midway, right, xshift=-0.3em, yshift=+0.3em] {\tiny +2.2}
([yshift=0.4em,xshift=0.em]{pic cs:7_dt_hota});
\draw[-latex,ForestGreen] ([yshift=0.3em,xshift=0.1em]{pic cs:6_dt_idf1})
to [out=-45,in=45]
node[midway, right, xshift=-0.3em, yshift=+0.3em] {\tiny +1.0}
([yshift=0.4em,xshift=0.em]{pic cs:7_dt_idf1});
\draw[-latex,ForestGreen] ([yshift=0.3em,xshift=0.1em]{pic cs:6_sm_hota})
to [out=-45,in=45]
node[midway, right, xshift=-0.3em, yshift=+0.3em] {\tiny +3.6}
([yshift=0.4em,xshift=0.em]{pic cs:7_sm_hota});
\draw[-latex,ForestGreen] ([yshift=0.3em,xshift=0.1em]{pic cs:6_sm_idf1})
to [out=-45,in=45]
node[midway, right, xshift=-0.3em, yshift=+0.3em] {\tiny +4.4}
([yshift=0.4em,xshift=0.em]{pic cs:7_sm_idf1});
\draw[-latex,ForestGreen] ([yshift=0.3em,xshift=0.1em]{pic cs:9_dt_hota})
to [out=-45,in=45]
node[midway, right, xshift=-0.3em, yshift=+0.3em] {\tiny +4.1}
([yshift=0.4em,xshift=0.em]{pic cs:10_dt_hota});
\draw[-latex,ForestGreen] ([yshift=0.3em,xshift=0.1em]{pic cs:9_dt_idf1})
to [out=-45,in=45]
node[midway, right, xshift=-0.3em, yshift=+0.3em] {\tiny +5.6}
([yshift=0.4em,xshift=0.em]{pic cs:10_dt_idf1});
\draw[-latex,ForestGreen] ([yshift=0.3em,xshift=0.1em]{pic cs:9_sm_hota})
to [out=-45,in=45]
node[midway, right, xshift=-0.3em, yshift=+0.3em] {\tiny +3.7}
([yshift=0.4em,xshift=0.em]{pic cs:10_sm_hota});
\draw[-latex,ForestGreen] ([yshift=0.3em,xshift=0.1em]{pic cs:9_sm_idf1})
to [out=-45,in=45]
node[midway, right, xshift=-0.3em, yshift=+0.3em] {\tiny +4.8}
([yshift=0.4em,xshift=0.em]{pic cs:10_sm_idf1});
\end{tikzpicture}
}
\caption{Ablation study on the validation set of each dataset. App stands for appearance embeddings, EMA for exponential moving average, Bb for bounding box, KF for Kalman Filter's predicted box, Kp for keypoints, and DA for the Data Augmentation.}
\label{tab:main_ablation_study}
\end{table}

\mysection{Temporal Encoders vs. Heuristics} These experiments compare our TE with standard heuristics using different cues. Regarding Re-ID features, TE consistently outperforms the Exponential Moving Average (EMA) (Exp. 1-2). This improvement is particularly noteworthy as appearance is a weak cue on DanceTrack but highly discriminative on SportsMOT. Similarly for bounding box cues, TE outperforms Kalman Filter's (KF) predictions on DanceTrack's erratic movements and frequent occlusions (Exp. 3-4). On the other hand, KF effectively captures the more predictable player trajectories in SportsMOT.
Pose keypoints provide complementary information, especially for distinguishing dancers during occlusions, but show no improvement over bounding box tracking on SportsMOT, likely due to noisy estimates from distant views (Exp. 5).

\mysection{Feature Fusion Analysis} We evaluate GAFFE's learned dynamic feature fusion against static rules. The baseline using equal weights for motion and appearance features (Exp. 6) shows no significant gain over using cues independently, and sometimes even decreases performance. Adding GAFFE for group-aware feature fusion (Exp. 7) yields consistent improvements, demonstrating the benefits of a learned approach. Using both temporal and group-aware encoding (Exp. 8) provides additional gain, with DanceTrack particularly benefiting from this combination.

\mysection{Complete Architecture and Training} Ablating data augmentation (Exp. 9) during our association-centric training significantly degrades performance, demonstrating the importance of training on diverse scenarios. Our final architecture with pose information (Exp. 10) achieves the strongest results on DanceTrack while showing no improvements on SportsMOT, likely due to its distant camera setup.

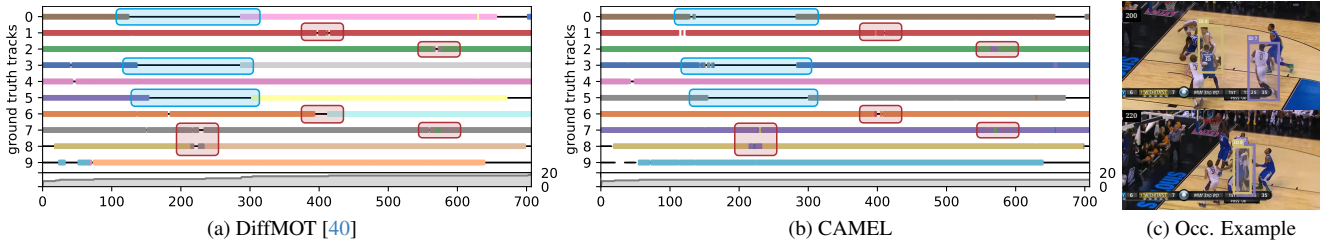
\begin{figure*}[ht]
    \centering
    \begin{subfigure}[b]{0.42\linewidth}
         \centering
         \begin{annotatedFigure}
        	{\includegraphics[width=1.0\textwidth]{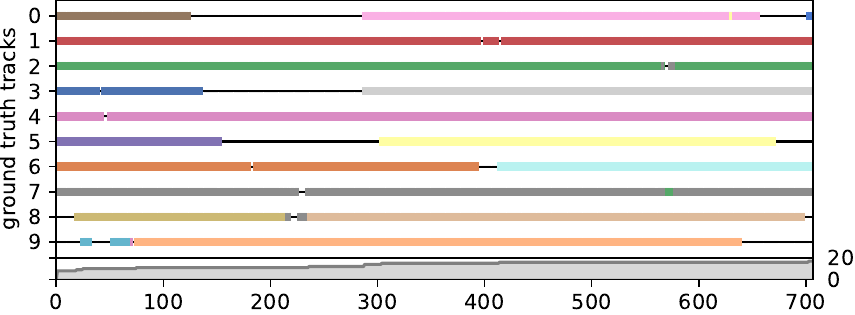}}
        	\annotatedFigureBox{0.20,0.9063}{0.46,0.9854}{SR}{0.20,0.9063}{Cerulean}
	\annotatedFigureBox{0.212,0.6647}{0.448,0.7438}{SR}{0.212,0.6647}{Cerulean}
	\annotatedFigureBox{0.227,0.5072}{0.459,0.5905}{SR}{0.227,0.5072}{Cerulean}
	\annotatedFigureBox{0.535,0.4257}{0.611,0.5107}{Occ}{0.535,0.4257}{Maroon}
	\annotatedFigureBox{0.309,0.2656}{0.385,0.424}{Occ}{0.309,0.2656}{Maroon}
	\annotatedFigureBox{0.746,0.7489}{0.822,0.8241}{Occ}{0.746,0.7489}{Maroon}
	\annotatedFigureBox{0.747,0.3524}{0.823,0.4274}{Occ}{0.747,0.3524}{Maroon}
    \annotatedFigureBox{0.535,0.8302}{0.611,0.9152}{Occ}{0.535,0.8302}{Maroon}

        \end{annotatedFigure}

         \caption{DiffMOT \cite{diffmot}}
     \end{subfigure}
     \hfill
     \begin{subfigure}[b]{0.42\linewidth}
         \centering
         \begin{annotatedFigure}
        	{\includegraphics[width=1.0\textwidth]{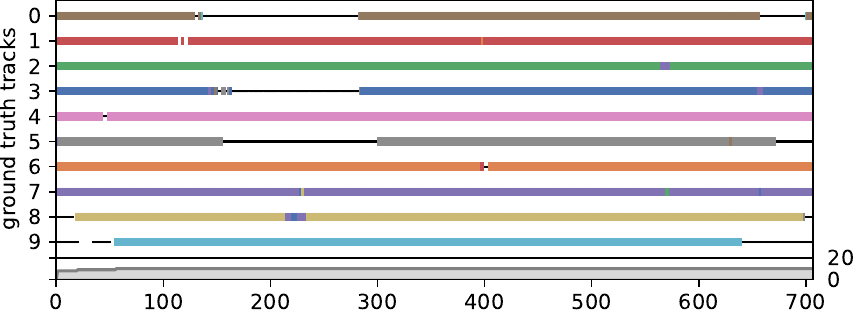}}
        	\annotatedFigureBox{0.20,0.9063}{0.46,0.9854}{SR}{0.20,0.9063}{Cerulean}
	\annotatedFigureBox{0.212,0.6647}{0.448,0.7438}{SR}{0.212,0.6647}{Cerulean}
	\annotatedFigureBox{0.227,0.5072}{0.459,0.5905}{SR}{0.227,0.5072}{Cerulean}
	\annotatedFigureBox{0.535,0.4257}{0.611,0.5107}{Occ}{0.535,0.4257}{Maroon}
	\annotatedFigureBox{0.309,0.2656}{0.385,0.424}{Occ}{0.309,0.2656}{Maroon}
	\annotatedFigureBox{0.746,0.7489}{0.822,0.8241}{Occ}{0.746,0.7489}{Maroon}
	\annotatedFigureBox{0.747,0.3524}{0.823,0.4274}{Occ}{0.747,0.3524}{Maroon}
    \annotatedFigureBox{0.535,0.8302}{0.611,0.9152}{Occ}{0.535,0.8302}{Maroon}
        \end{annotatedFigure}
         \caption{CAMEL}
     \end{subfigure}
     \hfill
     \begin{subfigure}[b]{0.15\linewidth}
         \centering
         \includegraphics[width=\textwidth]{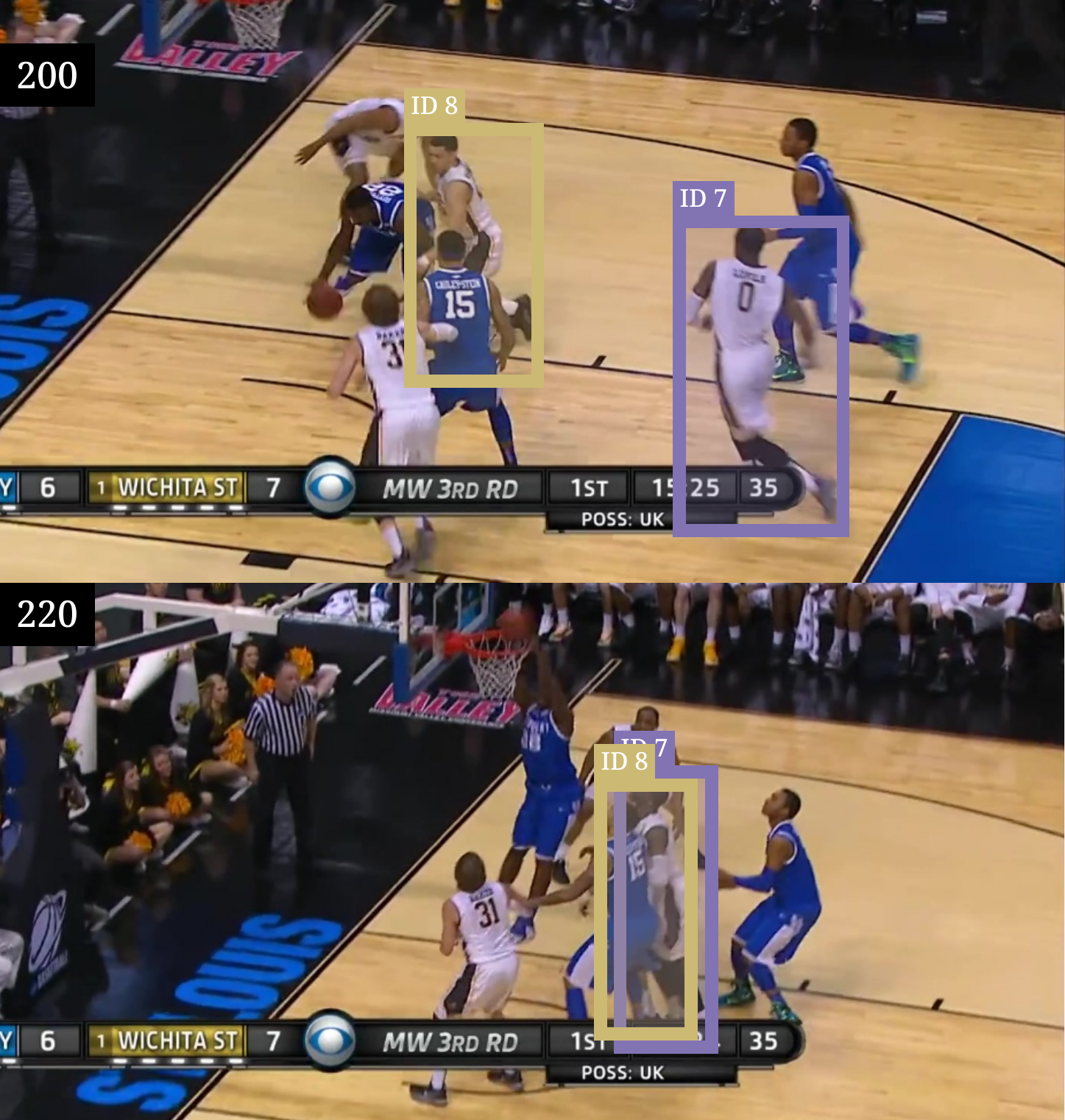}
         \caption{Occ. Example}
     \end{subfigure}
    \caption{Visualization of tracking results on \textit{v\_00HRwkvvjtQ\_c007} from SportsMOT. Ground truth tracks are depicted with horizontal lines, while colors indicates predicted identities. \textcolor{Cerulean}{Blue} zones highlight scene re-entries and \textcolor{Maroon}{red} zones show occlusions. Frames where the ground truth identity has left the scene are represented with a black line, and missing predictions are left blank. Bottom gray plots show cumulative tracked identities over time. (c) Frames from the highlighted occlusion between ids 7 \& 8 around frame 200.}
    \label{fig:qualitative-results}
\end{figure*}

\mysection{Analyzing TbD Association through Oracles} \label{subsec:oracle}
Two oracle experiments, detailed in \cref{sec:oracle_study}, have been designed to study the limitations of Tracking-by-Detections (TbD) heuristic-based association and evaluate the discriminative power of motion and appearance cues.
First, we design a Feature Fusion Oracle (Exp. 11) that linearly combines motion and appearance cues so as to result in a cost matrix that maximizes the association accuracy.
This oracle reveals two key insights: (i) motion and appearance are two strong and highly complementary cues for tracking, but (ii) the significant gap with standard fusion methods (Exp. 6) reveals that static heuristics fail to fully leverage their discriminative power.
Second, the Association Oracle (Exp. 12), which matches each detection to its IoU-closest ground truth track, establishes an absolute upper bound on association performance with detection quality as the only limitation.
The performance gap between Feature Fusion and Association Oracles varies significantly across datasets: the small gap on SportsMOT indicates reliable tracking cues, while the large gap on DanceTrack reveals the need for stronger cues in such challenging scenarios.
Overall, we find encouraging the results showing that \textit{our learned association strategy contributes to bridge the gap towards oracle performance} (Exp. 10-11 achieve close performances).

\subsection{Qualitative Analysis of Latent Representations} 
To illustrate CAMEL's cue disentanglement capabilities, we analyze similarity distributions between tracklet-detection pairs and latent space structure using t-SNE~\cite{tsne}. We compare CAMEL's output embeddings with standard heuristic cues: Kalman Filter (KF) for motion and Exponential Moving Average (EMA) of Re-ID embeddings for appearance.

\definecolor{posgreen}{RGB}{125,186,145}
\definecolor{negred}{RGB}{209,57,67}

\begin{figure}[hb]
    \centering
     \begin{subfigure}[b]{0.326\linewidth}
         \centering
         \begin{annotatedFigure}
        	{\includegraphics[width=\textwidth]{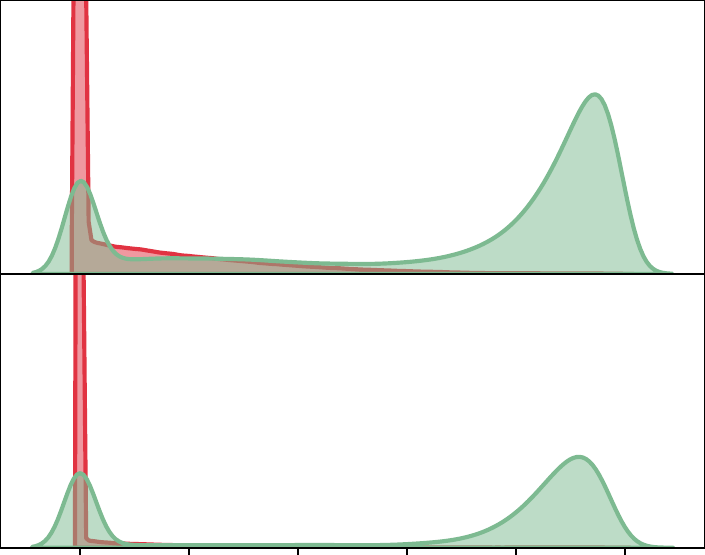}}
        	\annotatedRotFigureText{-0.013,0.55}{black}{0.2464}{\tiny DanceTrack}
            \annotatedRotFigureText{0,0.05}{black}{0.2464}{\tiny SportsMOT}
        \end{annotatedFigure}
         \caption{Motion}
     \end{subfigure}
     \hfill
     \begin{subfigure}[b]{0.326\linewidth}
         \centering
         \hspace*{0.2pt}
         \begin{annotatedFigure}
        	{\includegraphics[width=\textwidth]{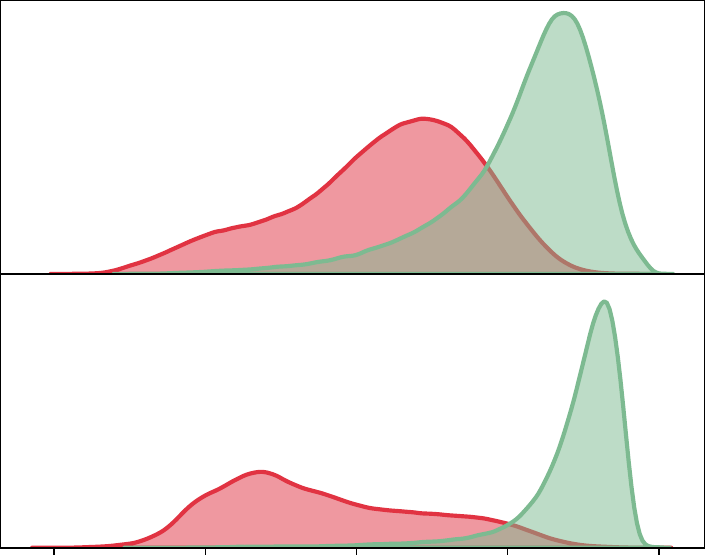}}
        \end{annotatedFigure}
         \caption{Appearance}
     \end{subfigure}
    \hfill 
     \begin{subfigure}[b]{0.326\linewidth}
         \centering
         \begin{annotatedFigure}
        	{\includegraphics[width=\textwidth]{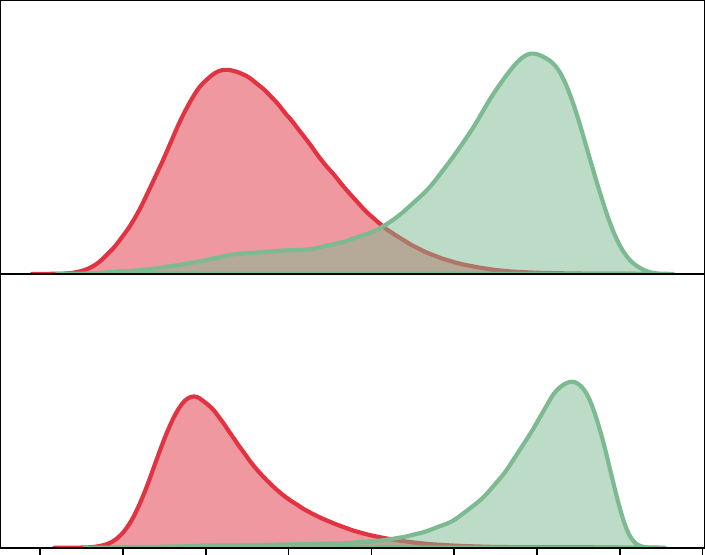}}
        \end{annotatedFigure}
         \caption{CAMEL output}
     \end{subfigure}
    \caption{
    Similarity distributions between \textcolor{posgreen}{positive} and \textcolor{negred}{negative} tracklet-detection pairs. (a) IoU between KF's predictions and detections. Cosine distances between : (b) EMA tracklet and detection ReID embeddings (c) pairs of CAMEL's output embeddings.}
    \label{fig:sample-distributions}
\end{figure}

\begin{figure}[ht]
    \centering
     \begin{subfigure}[b]{0.326\linewidth}
         \centering
         \includegraphics[width=\textwidth]{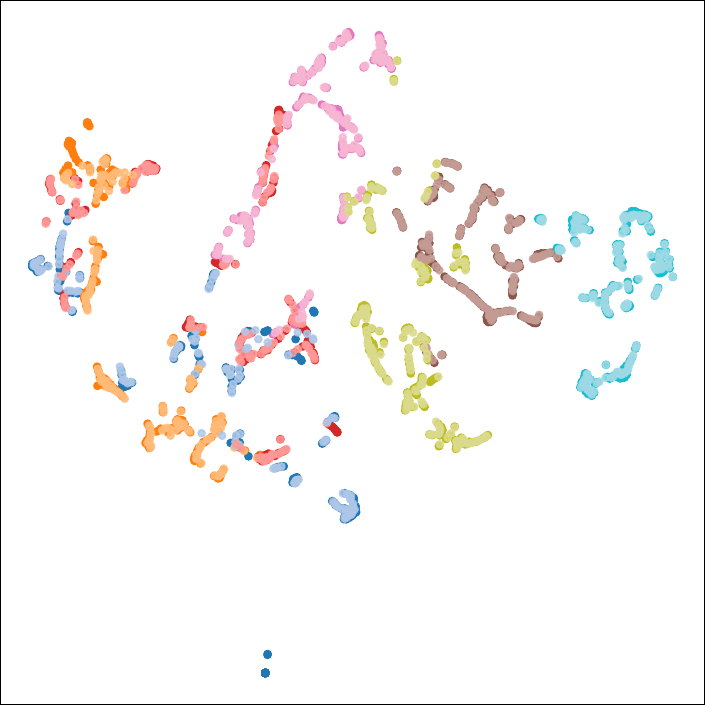}
         \caption{Motion}
     \end{subfigure}
     \hfill
     \begin{subfigure}[b]{0.326\linewidth}
         \centering
         \includegraphics[width=\textwidth]{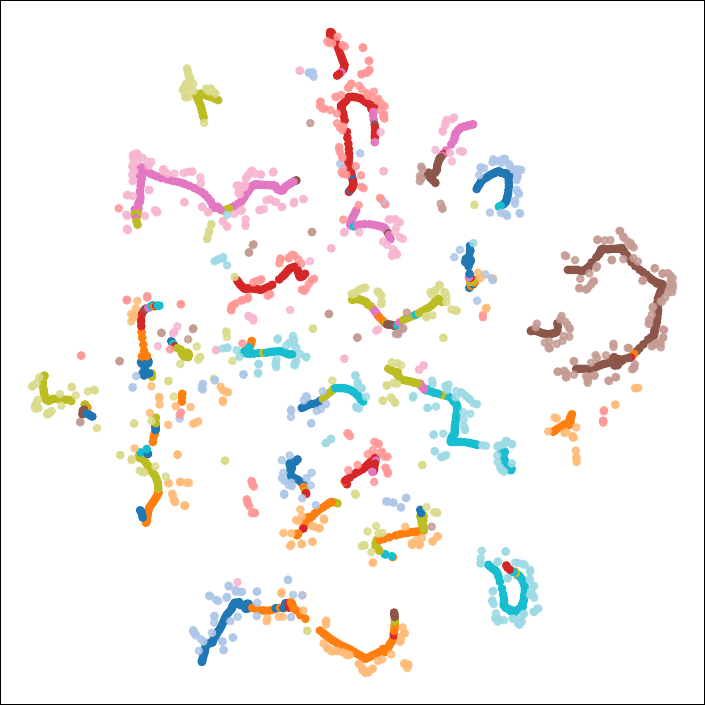}
         \caption{Appearance}
     \end{subfigure}
     \hfill
     \begin{subfigure}[b]{0.326\linewidth}
         \centering
         \includegraphics[width=\textwidth]{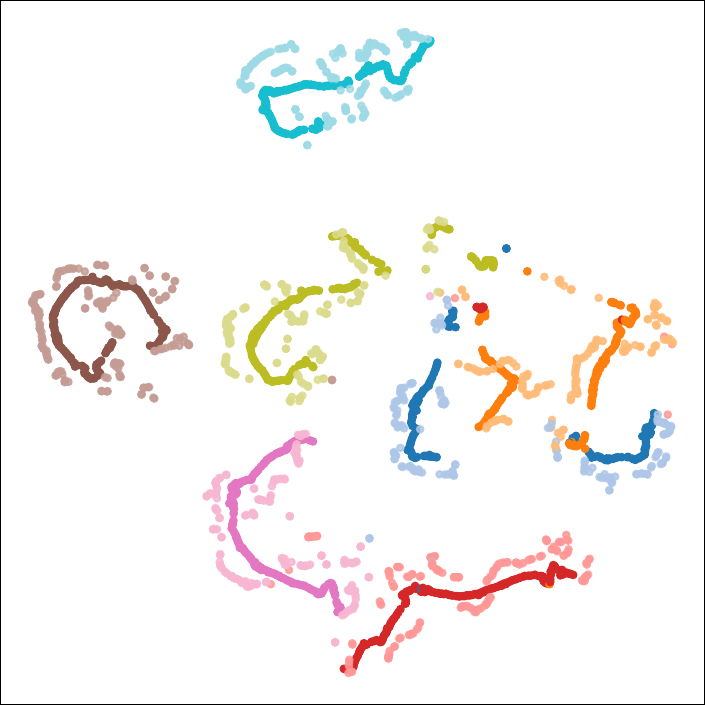}
         \caption{CAMEL output}
     \end{subfigure}
    \caption{t-SNE on the first 150 frames of \textit{dancetrack0019}. Each identity is assigned a unique color, with light/dark shades indicating detections/tracklets respectively. 
    }
    \label{fig:tsne}
\end{figure}

\mysection{Analysis of Similarity Distributions} 
\cref{fig:sample-distributions} compares the similarity distributions between tracklet-detection pairs that share the same identity (\textcolor{posgreen}{positive}) versus pairs with different identities (\textcolor{negred}{negative}), for standard motion/appearance cues and CAMEL's outputs.
While KF motion cues effectively discriminates most positive pairs from negative ones, a significant portion still exhibits incorrect low IoU values.
This limitation is particularly evident on DanceTrack, where negative pairs frequently overlap with positive ones, highlighting KF's weakness.
Moreover, appearance alone lacks discriminativeness, as evidenced by the non-negligible overlap between positive and negative pairs, especially on DanceTrack.
In contrast, CAMEL's output embeddings effectively discriminate positive from negative pairs, demonstrating a successful cue disentanglement.

\mysection{Latent Space analysis through t-SNE}
\cref{fig:tsne} illustrates the t-SNE representations of motion, appearance, and CAMEL outputs on a short sequence with heavy occlusions.
Motion embeddings organize into identity clusters but show significant overlap during occlusions, while appearance features achieve better but incomplete separation. On the other hand, CAMEL outputs form distinct identity clusters with minimal overlap, demonstrating an effective combination and disentanglement of these complementary cues.

\subsection{Qualitative Results}
\cref{fig:qualitative-results} compares CAMELTrack with the competitive DiffMOT~\cite{diffmot} using the same detections on a challenging SportsMOT sequence featuring scene re-entries and heavy occlusions. This figure illustrates their tracking performance through a timeline where ground truth tracks are represented with horizontal lines, and identities with different colors. For both methods, a cumulative plot shows the total number of unique identities created over time.

Both methods show distinct behaviors during scene re-entries: while DiffMOT generates new identities, CAMEL successfully recovers known ones through its feature bank, as shown by the lower slope in the cumulative identity plot. Similarly, during occlusions, both methods initially make identity switches, but CAMEL recovers from these errors while DiffMOT propagates them forward.

\section{Conclusion}

We introduced CAMEL, a novel learned association module that replaces traditional hand-crafted rules—tracklet representation, feature fusion, and multi-stage matching—with a unified trainable architecture. 
With our state-of-the-art performance, we view this work as a first step to reaffirm TbD as a strong paradigm for online tracking and encourage a shift from association heuristics toward fully learned approaches. We release our code to foster future research in this direction.
Building upon CAMEL, future work could explore more sophisticated training objectives and neural architectures, or extend the learning paradigm to other components like tracklet life cycle management.

\paragraph{Acknowledgments.}
This work has been funded by Sportradar and by the project ReconnAIssance of Pole MecaTech - Region Wallonne. Computational resources have been provided by the Consortium des Équipements de Calcul Intensif (CÉCI), funded by the Fonds de la Recherche Scientifique de Belgique (F.R.S.-FNRS) under Grant No. 2.5020.11 and by the Walloon Region.

{
    \small
    \bibliographystyle{ieeenat_fullname}
    \bibliography{bib/abbreviation-short, bib/formatted_bib}
}


\appendix
\clearpage
\setcounter{page}{1}
\maketitlesupplementary

\noindent The supplementary material includes the following sections.
\begin{itemize}
    \item A methodological comparison highlighting the key architectural differences with previous transformer-based trackers (\cref{sec:comparison_architecture}).
    \item Some extended results on PoseTrack21 and BEE24 benchmarks (\cref{sec:extended_results}).
    \item A detailed discussion of MOT17 limitations and comparison with state-of-the-art methods (\cref{sec:mot17_discussion}).
    \item The extended implementation details covering detection, pose, re-identification models and life cycle management (\cref{sec:extended_implementation_details}).
    \item The training procedure details, including preprocessing steps, training loop specifics, and data augmentation strategies (\cref{sec:detailed_training}).
    \item More details about the Oracles implemented for the ablation study (\cref{sec:oracle_study}).
    \item An extended related work discussion with additional comparisons (\cref{sec:detailed_related_work}).
    \item Some additional qualitative results (\cref{sec:additional_qualitative_results}).
\end{itemize}

\noindent Our GitHub repository is available here: \url{https://github.com/TrackingLaboratory/CAMELTrack}

\section{Methodological Differences Compared to Previous Transformer-based Trackers} \label{sec:comparison_architecture}
The core objective of our work was to integrate a fully-learned, heuristic-free association module in online tracking-by-detection (TbD). Surprisingly, the research direction has been unexplored in previous works, with researchers favoring the E2E paradigm despite its many drawbacks discussed in the Related Work (\cref{sec:related_work}). CAMELTrack demonstrates that heuristics can be completely replaced with an simple and elegant solution, that, despite its apparent simplicity, \uline{differs fundamentally from existing transformer-based tracking methods}. We detail these key differences in this section.

\subsection{Comparison to TransMOT and TransTAM}
CAMELTrack differs significantly from previous tracking-by-detection transformer-based methods like TransMOT~\cite{transmot} and TransTAM~\cite{transtam}. These differences span multiple aspects of the architectural design:
\begin{enumerate}
\item \textbf{N×M Edge Tokens vs N+M Object Tokens:} TransMOT and TransTAM adopt graph-inspired approaches using N×M edge tokens (one edge token for each tracklet-detection pair) and employ token-wise binary classifiers to predict an assocation score for each token and thereby approximate the tracklet-detection association matrix. CAMEL follows a fundamentally different approach based on deep metric learning. Our method encodes both tracklets and detections (N+M tokens) into a shared disentangled latent space where associations are determined through cosine distance comparisons. This architectural distinction not only simplifies the design but also proves more effective, eliminating the quadratic complexity of edge-based representations.
\item \textbf{Explicit Attention Bias:} TransMOT and TransTAM both introduce an explicit spatial bias into the attention mechanism of their Spatial Graph Transformer, artificially restricting communication between spatially adjacent detections having a non-null IoU. We found that such explicit spatial bias is unnecessary in our architecture and that global tracklet-detection communication through unbiased self-attention yields superior results, thereby demonstrating the improved learning capabilities of our design.
\item \textbf{Cue Fusion:} TransMOT naively concatenates a re-identification embedding with a vector of 4 scalars representing the bounding box. On the other hand, CAMEL first embeds each cue independently into a higher-dimensional space with an end-to-end learnable FFN before summing them. CAMEL also performs independent temporal encoding per cue before fusion, whereas TransMOT and TransTAM perform fusion first, followed by temporal encoding. This strategy allows our network to treat each cue in an agnostic and balanced way, ensuring any type of cue can be easily added to the system while maintaining equal importance between different cues to achieve better feature fusion.
\item \textbf{Association-centric Training (ACT):} TransMOT trains on short time windows reducing their potential to solve long-term tracking, does not perform data augmentation on the tracklet-detection pairs, and do not build synthetic association training examples combining multiple videos, as our ACT does. 
\item \textbf{Heuristic Dependency:} First, TransMOT is not heuristic-free. Indeed, it relies on a hand-crafted multi-stage matching pipeline where their learned transformer module is only used in the second stage, while the first and third stages remain purely based on heuristics, with a bounding box IoU and ReID cosine distance for the first and third stages respectively. In contrast, CAMELTrack is fully heuristic-free, with a single association stage.
\end{enumerate}
TransTAM, while attempting a heuristic-free approach, is an unpublished arXiv work that performs worse than TransMOT despite incorporating offline post-processing techniques. Our CAMELTrack maintains a fully online architecture yet outperforms both methods by approximately 2\% MOTA on MOT17. Unfortunately, TransMOT does not provide HOTA performance metrics and has no publicly available code for further comparative analysis.

\subsection{Comparison to DETR-based E2E Methods}
The vast majority of previous transformer-based trackers follow the end-to-end (E2E) paradigm with architectures based on DETR. These methods, exemplified by MOTR~\cite{motr}, differ fundamentally from CAMEL in several key aspects. First, DETR-based methods employ a transformer decoder that processes track and detection queries through cross-attention to CNN feature maps, performing object detection with implicit association. In contrast, CAMEL utilizes a transformer encoder that processes high-level tracklet and detection tokens through self-attention, with no reliance on low-level CNN feature maps. Second, while DETR-based approaches handle association implicitly within the detection process, CAMEL solves association explicitly using the Hungarian algorithm on the encoded tokens. The DETR transformer must perform both detection and association jointly, which creates a drawback on performance due to antagonist objectives, as discussed extensively in previous works \cite{motrv2, motip}. Finally, CAMEL's architecture enables the processing and fusion of various input cues from off-the-shelf expert models, whereas DETR-based methods depend on object re-detection within CNN feature maps.

\section{Extended results on PoseTrack21 and BEE24}

\begin{table}[t]
    \centering
    \renewcommand{\arraystretch}{0.9} 
    \resizebox{\columnwidth}{!}{
    \begin{tabular}{l*{5}{c}}\toprule
        Method                      & HOTA$\uparrow$ & AssA$\uparrow$ & DetA$\uparrow$ & IDF1$\uparrow$ & MOTA$\uparrow$ \\\midrule
        \hspace{-0.5em}{\scriptsize \color{gray} Public Detection Setting} \\
        CorrTrack \cite{posetrack21}& 57.0 & 64.2 & 51.3 & 66.5 & 52.0 \\
        GAT \cite{gat}              & 58.4 & 66.9 & \textbf{51.8} & - & \textbf{55.3} \\
        \cmidrule(lr){1-6}
        CAMELTrack                       & \textbf{58.7} & \textbf{70.7} & 50.0 & \textbf{67.8} & 51.7 \\
        \hline
        \hspace{-0.5em}{\scriptsize \color{gray} Private Detection Setting} \\
        TRMOT \cite{TRMOT}          & 46.9 & 55.0 & 40.9 & 57.3 & 47.2 \\
        FairMOT \cite{fairmot}      & 53.5 & 61.5 & 47.4 & 63.2 & 56.3 \\
        Tracktor++ \cite{tracktor}  & 58.3 & 65.4 & 52.7 & 69.3 & 59.5 \\
        \cmidrule(lr){1-6}
        CAMELTrack                       & \textbf{66.0} & \textbf{73.8} & \textbf{59.9} & \textbf{76.0} & \textbf{67.5} \\
        \bottomrule
    \end{tabular}
    }
    \caption{Comparison on PoseTrack21~\cite{posetrack21} validation set.}
    \label{tab:posetrack_sota}
\end{table}

\begin{table}[t]
    \centering
    \renewcommand{\arraystretch}{0.9} 
    \resizebox{0.9\columnwidth}{!}{
    \begin{tabular}{lccHcc}\toprule
        Method          & HOTA$\uparrow$ & AssA$\uparrow$ & DetA$\uparrow$ & IDF1$\uparrow$ & MOTA$\uparrow$ \\\midrule
        \hspace{-0.5em}{\scriptsize \color{gray} End-to-End MOT} \\
        TrackFormer \cite{trackformer} & 44.3 & 42.3 & - & 53.9 & 41.5 \\
        
        \hline
        \hspace{-0.5em}{\scriptsize \color{gray} Heuristic Association} \\
        \rowcolor{babyblue!20}
        ByteTrack \cite{bytetrack} & 43.2 & 38.3 & - & 56.8 & 59.2 \\
        \rowcolor{babyblue!20}
        OC-SORT \cite{ocsort} & 42.7 & 36.8 & - & 55.3 & 61.6 \\
        \rowcolor{babyblue!20}        
        TOPICTrack \cite{bee24} & 46.6 & 40.3 & - & 59.7 & 66.7 \\
        \hline
        \hspace{-0.5em}{\scriptsize \color{gray} Learned Association} \\
        \rowcolor{babyblue!20}
        CAMELTrack           & \textbf{50.3} & \textbf{42.6} & 59.6 & \textbf{63.8} & \textbf{75.7} \\
        \bottomrule
    \end{tabular}
    }
    \caption{Comparison on BEE24~\cite{bee24} test set. Methods with {\color{babyblue} blue} background use the same YOLOX detector.}
    \label{tab:bee24_sota}
\end{table}

\label{sec:extended_results}
\mysection{PoseTrack21} PoseTrack21~\cite{posetrack21} serves as a diverse real-world testbed where we demonstrate our method's modularity through effective keypoint integration.

As reported in~\cref{tab:posetrack_sota}, current methods can be categorized into two settings: methods with private detections extending traditional MOT frameworks~\cite{TRMOT,fairmot,tracktor}, and those using public detections with a custom pose-aware tracking~\cite{gat,posetrack21}.

Different from SportsMOT and DanceTrack, PoseTrack covers diverse real-world scenarios with dramatic camera motion, viewpoint changes, and motion blur, making detection and association challenging.

Using public detections from~\cite{posetrack21,gat}, CAMEL establishes new state-of-the-art performance with significant gains ($+3.8\%$ AssA) through an effective fusion of appearance, motion and pose cues.
With stronger private detections, CAMEL achieves even larger gains ($+7.7\%$ HOTA).

\mysection{BEE24} BEE24~\cite{bee24} is a novel MOT benchmark showcasing complex motion, heavy occlusions, difficult re-identification, and long sequences (up to 5000 frames). As demonstrated in \cref{tab:bee24_sota}, CAMELTrack surpasses existing state-of-the-art methods by at least +3.7\% HOTA. In particular TOPICTrack~\cite{bee24}
, which employs specialized heuristics designed to model the intricate dynamics between rapid bee flight motion and heavy occlusions on the ground.

The BEE24 experiments confirm CAMELTrack's effective transferability to new domains with minimal adaptation requirements. Our implementation utilizes only bounding box positional data, as re-identification models proved inadequate for distinguishing individual bees. This positional-only approach highlights our framework's flexibility for incremental deployment, where additional cues can be incorporated when available but are not essential for robust performance.

\section{MOT17 Discussion} \label{sec:mot17_discussion}

MOT17 remains a well-established dataset within the MOTChallenge benchmark and has historically served as a primary evaluation platform for multi-object tracking. However, as highlighted by recent works \cite{transmot, motrv2, motip, memotr, motr}, several fundamental limitations make MOT17 particularly unsuitable for evaluating learned association approaches like CAMEL. These limitations, combined with our comprehensive evaluation on SportsMOT, DanceTrack, and PoseTrack21, motivate our choice to present MOT17 results in the supplementary material. For completeness, we discuss important limitations of MOT17 when evaluating learned-based tracking methods, before providing our test results and comparison with state-of-the-art methods.

\subsection{MOT17 Dataset Limitations}

MOT17 consists of 7 training sequences, totaling approximately 5.9K frames (215 seconds of video), with a test set of 7 others videos whose annotations are kept private. Results must be submitted through an official evaluation server that enforces a 72-hour waiting period between submissions and a maximum of 4 submissions per method.

A critical limitation of MOT17 (and similarly, MOT20) is the \textbf{absence of a proper validation set}, which severely impedes the development and evaluation of learned MOT approaches, like End-To-End (E2E) methods or CAMEL. Popular works~\cite{transmot, motrv2, motip, ghost, bytetrack, ocsort} commonly create a validation split using the second half of all training sequences. However, we believe this practice is methodologically unsound, especially for learned methods, as it is prone to overfitting since both portions share the same scene characteristics and often contain the same tracked identities. This lack of proper validation set prevents meaningful ablation studies and proper model validation. 

MOT17's dataset design thus \textbf{inherently favors heuristic-based methods} that require only hyperparameter optimization, over data-driven approaches that need a proper training and validation set. This bias is reflected in the benchmark's leaderboard, which is dominated by heuristic trackers. On the other hand, learned approaches typically underperform on MOT17 despite their success on other datasets.

\subsection{Comparison with state-of-the-art methods}
Results on the MOT17 test set are reported in \cref{tab:mot17_sota}. Despite the limitations discussed above, CAMEL outperforms all existing learned approaches and achieves competitive performance with state-of-the-art heuristic methods. The comparison with each type of method in the prior art is detailed below.

\mysection{End-to-End MOT}
As discussed in \textbf{MOTIP}~\cite{motip}, end-to-end (E2E) approaches, which jointly learn detection and association, require substantial training data. Most of these methods leverage the CrowdHuman~\cite{crowdhuman} dataset for joint training to overcome this limitation. Despite not utilizing additional training data, our method still outperforms these E2E approaches.

\mysection{Hybrid Association}
As detailed in \cref{sec:detailed_related_work}, \textbf{TADN}~\cite{tadn} and \textbf{TransMOT}~\cite{transmot} represent initial attempts to integrate learned components into Tracking-by-Detection (TbD) pipelines. However, our approach outperforms both methods, which we attribute to two main factors: our Association-Centric Training addresses key limitations in their training strategies by (i) overcoming their reliance on short training sequences (e.g., 5 frames for TransMOT), and (ii) leveraging data augmentation to produce rich and diverse training samples. Additionally, our method likely benefits from more discriminative appearance cues. TransMOT adopts a conservative tracking strategy that prioritizes identity preservation, resulting in high IDF1 scores but lower MOTA due to its tendency to miss detections (high false negatives).

\mysection{Heuristic Association}
SORT-based methods, which rely on handcrafted association rules, have long dominated the MOTChallenge benchmark. As discussed in the previous section, the structure of the dataset inherently favors such methods, as they require only a small training set to optimize their hyperparameters. Despite MOT17's inherent bias towards such methods, our learned CAMELTrack achieves competitive performance.

\mysection{About Online Methods Using Offline Post-Processing}
For a fair comparison with our fully online CAMELTrack, we report performance only against other fully online methods in \cref{tab:mot17_sota}. We exclude several state-of-the-art methods~\cite{botsort, diffmot, hybridsort, strongsort, deepocsort} that, despite their online nature, employ offline post-processing mechanisms to boost their performance on MOT17.
Specifically, \textbf{ByteTrack}~\cite{bytetrack} utilizes sequence-specific detection thresholds combined with linear interpolation, as thoroughly analyzed in GHOST~\cite{ghost}. We however report the performance reported in GHOST~\cite{ghost}, that ran ByteTrack with a single threshold and without interpolation. While \textbf{DiffMOT}'s~\cite{diffmot} official paper and repository do not explicitly mention interpolation, a careful investigation of their official results\footnote{\url{https://github.com/Kroery/DiffMOT/releases/download/v1.2/MOT17_DiffMOT.zip}} reveals the use of such techniques: despite using the same YOLOX detections as ByteTrack (which are widely adopted by popular methods), their submission contains additional detections that are characteristic of interpolation. \textbf{StrongSORT}~\cite{strongsort} incorporates two offline post-processing modules as detailed in their paper: Appearance Free Link (AFLink) and Gaussian-Smoothed Interpolation (GSI). Similarly, \textbf{Hybrid-SORT}\footnote{\url{https://github.com/ymzis69/HybridSORT}}~\cite{hybridsort} and \textbf{BoT-SORT}\footnote{\url{https://github.com/NirAharon/BoT-SORT}}~\cite{botsort} employs interpolation techniques as documented in their official GitHub repositories.

We opt not to report performance with interpolation for two main reasons: (i) our focus on developing truly online tracking solutions, and (ii) the submission limitations on the official MOTChallenge evaluation server discussed above.

\section{Extended Implementation Details} \label{sec:extended_implementation_details}
Our complete implementation, including configuration files, model weights and used detections, is publicly available at \url{https://github.com/TrackingLaboratory/CAMELTrack}. We encourage readers to refer to our codebase for full methodological details and reproducibility.

\mysection{Detections}
For fair comparison, we use the same detection setup as DiffMOT~\cite{diffmot}. Specifically, we employ YOLOX-x~\cite{yolox} models trained following ByteTrack's~\cite{bytetrack} procedure: for DanceTrack, we use weights provided by the original benchmark~\cite{dancetrack}; for SportsMOT, we use weights provided by MixSORT~\cite{sportsmot}; for MOT17, we directly use weights provided by ByteTrack. For PoseTrack21, we fine-tune a YOLOX-x model using MMDetection's~\cite{mmdetection} methodology. To encourage research focused on association rather than detection quality, we provide all detection results in the standardized MOT format.

\mysection{Pose Models}
For pose estimation, we leverage pre-trained models from MMPose~\cite{mmpose}: RTMPose~\cite{rtmpose} for DanceTrack and SportsMOT, and HRNet~\cite{hrnet} trained on PoseTrack18 for MOT17. For PoseTrack21, we fine-tune an HRFormer~\cite{hrformer} model following the PoseTrack18 training protocol but on PoseTrack21.

\mysection{Re-ID Models}
Similar to previous state-of-the-art SORT-based works \cite{diffmot, ghost, gat, posetrack21, deepeiou, sportsmot, hybridsort, deepocsort} that train their custom re-identification model for each dataset, we train our own re-identification model based on BPBReID \cite{bpbreid} to produce appearance cues.
Comparing methods without taking into account the performance of their ReID module is an impossible task, since some online TbD work don't use appearance cues \cite{ocsort, bytetrack}, other works like E2E methods learn appearance cues implicitly from their detection backbone \cite{motr, memotr, motip}, and most TbD pipelines employing a ReID module all trained their own custom trained model \cite{diffmot, ghost, gat, posetrack21, deepeiou, sportsmot, hybridsort, deepocsort}.

On DanceTrack~\cite{dancetrack}, DiffMOT~\cite{diffmot} employs the ReID model introduced by Deep-OCSORT~\cite{deepocsort}, GHOST~\cite{ghost} has its own model with test time domain adaptation, and Hybrid-SORT~\cite{hybridsort} trains a custom model jointly on DanceTrack and CUHKSYSU~\cite{cuhk}.
On SportsMOT, DiffMOT~\cite{diffmot} trains its own model on FastReID~\cite{fastreid}, Deep-EIoU~\cite{deepeiou} trains a custom OSNet~\cite{osnet} model and MixSORT introduces a novel appearance model.
On PoseTrack21~\cite{posetrack21}, CorrTrack-ReID~\cite{posetrack21} and GAT~\cite{gat} both have their custom appearance model.
Finally on MOT17~\cite{mot17}, DiffMOT~\cite{diffmot} and Hybrid-SORT~\cite{hybridsort} employ the ReID model provided by BoT-SORT, while GHOST trains a custom ResNet50-based model jointly on MOT17~\cite{mot17} and Market1501~\cite{market1501}, and MixSORT~\cite{sportsmot} uses again its custom appearance model.

As introduced in \cref{sec:implementation_details}, we train one BPBReID~\cite{bpbreid} model per dataset. BPBReID is a part-based ReID model that produces one embedding per body part, to increase robustness against occlusions. We first build a re-identification dataset from each of the train sets of the MOT datasets, by randomly picking up to 1000 tracklets, and then uniformly sampling along the temporal dimension up to 20 images per tracklet. We also build a validation set, using all tracklets from the corresponding MOT validation set, then sampling along the temporal dimension up to 10 images per tracklet. We then train BPBReID on these ReID datasets using the same recipe as the original paper, with 5 body parts and with a Swin \cite{swin} transformer backbone from the SOLIDER person foundation model \cite{solider}.

Our final ReID models achieve 81.8\%mAP on SportsMOT, 34.4\%mAP on DanceTrack and 84.9\%mAP on PoseTrack21. Performance on SportsMOT and PoseTrack21 are below what state-of-the-art models can achieve on the popular Market-1501 dataset \cite{market1501} (i.e. over 90\% mAP), highlighting the difficulty of re-identification in these domains, because of the similar appearance of multiple identities that share the same sport jersey. Moreover, on DanceTrack, we come to conclusions similar to those of the original paper~\cite{dancetrack}, with very low ReID performance. The previous tracking methods mentioned above \cite{ghost, botsort, diffmot, gat} don't disclose the raw ReID performance of their custom ReID model, rendering a comparison to our own model difficult.

Finally, when training CAMEL, we need to avoid using "perfect" (overfit) ReID embeddings, due to the fact that the ReID model producing these embeddings has been trained on that same data. To avoid such issue, we generate the training ReID embeddings of our Association-Centric Training set introduced in \cref{subsec:train} as follows. First, we train a ReID model on the first half of the training set, then use it to generate the ReID embeddings of the second half. We then do the opposite to generate the ReID embeddings of the first half. The ReID embeddings for the validation set are generated with a model trained on the full train set. The ReID embeddings for the test set are generated with a model jointly trained on the train and val set, similar to previous work \cite{hybridsort, deepeiou, diffmot}.

\mysection{Life Cycle Management}
Different parameters are used across datasets.
\begin{itemize}
    \item Detection confidence threshold: 0.4 (DanceTrack), 0.1 (SportsMOT), 0.3 (PoseTrack21), 0.5 (MOT17).
    \item Minimal confidence for tracklet initialization threshold: 0.9, 0.4, 0.4, 0.55 respectively.
    \item CAMEL's tracklet-detection similarity threshold: 0.1, 0.1, 0.45, 0.5 respectively.
\end{itemize}

Given reliable detections, minimum hits for tracklet confirmation is set to 0 for all datasets except MOT17, where we require 1 hit to filter sporadic detector noise.

\section{Detailed Training} \label{sec:detailed_training}

We supplement here the information in \cref{subsec:train}. Interested readers are encouraged to look at the code for the exact training procedure.
\subsection{Preprocessing}
We create our training dataset by combining the ground truth detection and tracklet identity information with cue-specific information from upstream models. 
\begin{enumerate}
    \item We perform Hungarian matching with IoU between the ground truth bounding boxes and a detector, in order to give each predicted bounding box a ground truth identity.
    \item Every resulting detection is then passed through every cue-specific model (reid, pose estimation).
    \item We compute the bbox overlap between detections in the same frame. This information is later required by some data augmentations.
    \item All resulting information is saved on disk.
\end{enumerate}

\subsection{Training Loop}
During training we sample from our pre-generated set to build training batches of $B$ training samples. One sample for training with $P$ tracklet-detection pairs is created through the following steps.
\begin{enumerate}
    \item Selecting a random frame from a random video.
    \item Gathering all detections from that frame and all detections from previous frames.
    \item Performing data augmentation on the tracklets and detections (see \cref{subsec:da}).
    \item Only keeping the $W$ last detections per tracklet (W=50 in most experiments).
    \item Repeating this procedure with a new frame until we obtain $P$ tracklet-detection pairs.
\end{enumerate}

CAMEL then receives all the detections and tracklets for the samples in a batch, and outputs one embedding for each detection and each tracklet. The InfoNCE~\cite{infonce} loss is then computed using the paired tracklet-detection embeddings, using the ground truth track ids to match each pair.


\subsection{Data Augmentations}\label{subsec:da}
We employ four different types of data augmentations. The augmentations are either fully random, or based on observed characteristics. The main detection characteristic we use is the IoU with other detections in the same frame.

\mysection{Detection Identity Swapping} To generate realistic identity switches, we randomly select a tracklet and find another tracklet that overlaps with it (i.e., both tracklets have at least one pair of detections with non-zero IoU). We then swap the identities of these overlapping detections to simulate tracking errors that typically occur during occlusions.

\mysection{Detection Dropout} This data augmentation removes detections within a tracklet with probability $p_{drop}$. To simulate challenging association scenarios like recovery after long occlusions, scene re-entries, we apply detection dropout with higher probability on more recent detections.


\mysection{Cue Dropout} We randomly remove specific cues (appearance, motion, or pose) from detections during training. Despite its intuitive appeal for improving robustness to missing cues, this augmentation showed no measurable impact on model performance.

\mysection{Random perturbations} Finally, we design a data augmentation that perturbs the input cues to improve model generalization. Specifically, we add Gaussian noise to appearance embeddings, bounding box coordinates, and keypoint coordinates.

The optimal parameters of each data augmentations are selected through a grid search on the validation set of each dataset.

\section{Oracle Study}

\label{sec:oracle_study}

We provide implementation details for the two oracle experiments referenced in our ablation study (\cref{subsec:ablation}).

\mysection{Association Oracle (Exp. 12)}
This oracle establishes an absolute upper bound on association performance, limited only by detection quality. During each association step of the online TbD pipeline :

\begin{enumerate}
    \item Current detections are matched to ground truth bounding boxes using the Hungarian algorithm;
    \item IoU score is used as the matching metric with a minimum threshold of 0.5;
    \item eEach matched detection inherits the track identity of its corresponding ground truth.
\end{enumerate}

\mysection{Feature Fusion Oracle (Exp. 11)}
This oracle demonstrates the potential of optimal feature fusion while highlighting current limitations of heuristic-based association rules. For each incoming frame, the following is applied.

\begin{enumerate}
    \item A single weight factor linearly combines appearance and motion costs into a unified cost matrix.
    \item The resulting cost matrix is processed by the Hungarian algorithm for final matching.
    \item The optimal weight is determined by maximizing the association accuracy (percentage of correct tracklet-detection matches), thanks to privileged access to ground-truth annotations.
\end{enumerate}

\mysection{Limitations and Future Extensions}
While our simple implementation sufficiently illustrates the limitations of current heuristic-based methods, more sophisticated oracles could be developed. For instance, computing per-tracklet optimal weights would better reflect how cue reliability varies across targets. This would be particularly relevant for scenarios where:

\begin{itemize}
    \item Appearance cues dominate for visually distinct targets (e.g., goalkeepers in soccer);
    \item Motion cues better discriminate between similarly-appearing targets (e.g., same-team players).
\end{itemize}

However, developing such advanced oracles extends beyond our current scope, as our simple oracle adequately demonstrates the potential for improvement in feature fusion strategies (see \cref{tab:main_ablation_study}).

\section{Detailed Related Work} \label{sec:detailed_related_work}

\begin{sidewaysfigure*}[p]
\centering
\includegraphics[width=\linewidth]{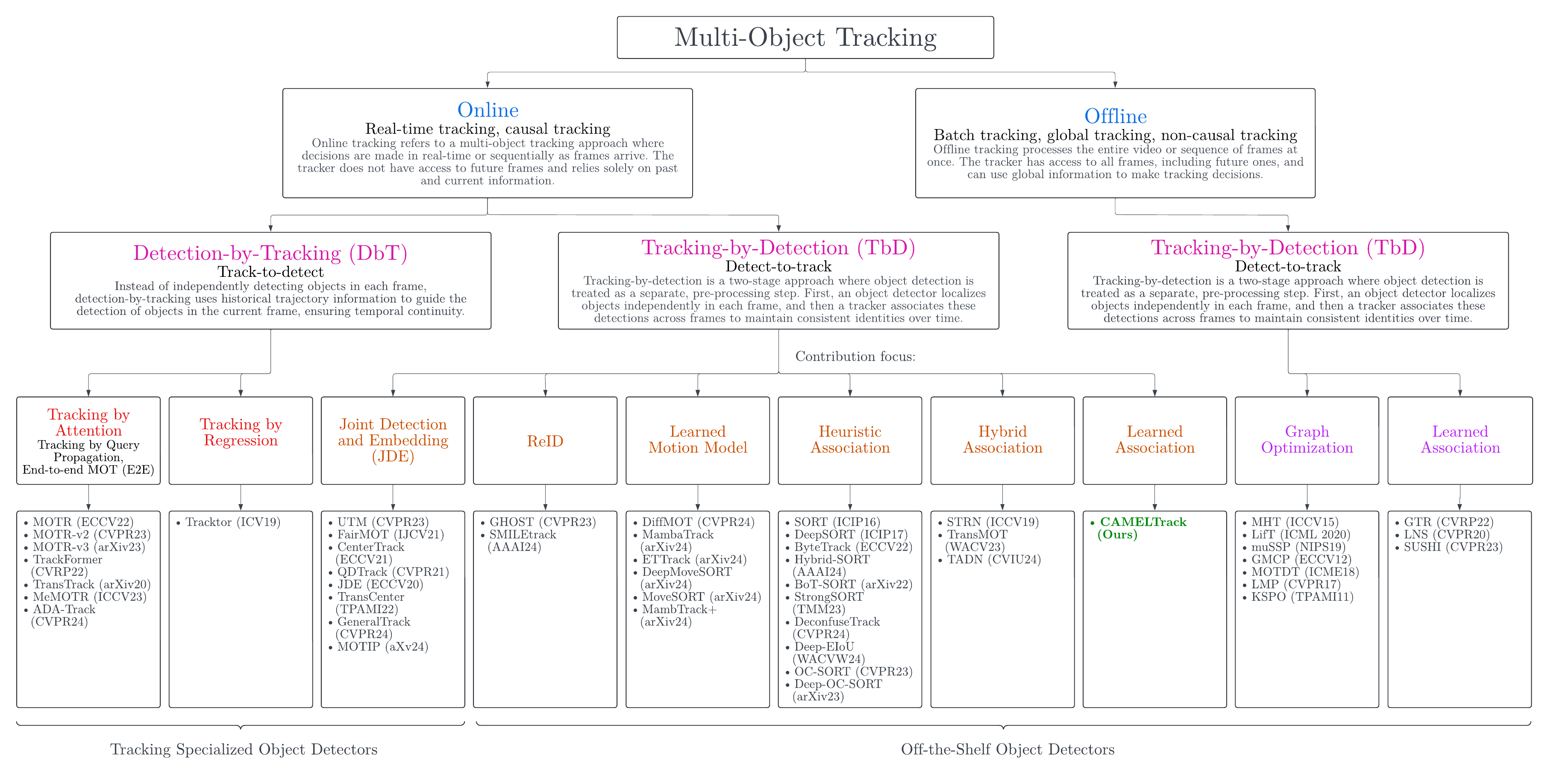}
  \caption{
    Taxonomy of current Multi-Object Tracking (MOT) approaches. CAMELTrack introduces a new direction by proposing a novel learned association module within the online tracking-by-detection paradigm.
  }
\label{fig:mot_sota_tree}
\end{sidewaysfigure*}

In this section, we complement \cref{sec:related_work} by providing a more comprehensive review of key Multi-Object Tracking (MOT) approaches related to our work, with particular focus on online methods.
\cref{fig:mot_sota_tree} illustrates the position of CAMELTrack in the current MOT taxonomy.

\mysection{Heuristic-based Tracking-by-Detection}

The dominant paradigm in MOT has been tracking-by-detection (TbD), with many methods building upon SORT~\cite{sort}. These approaches focus on developing sophisticated association heuristics~\cite{deepsort, bytetrack, strongsort, botsort}, or stronger motion modeling \cite{ocsort, diffmot, motiontrack, ettrack, deepmovesort, movesort, mambatrack, mambtrack} and re-identification~\cite{ghost, smiletrack, finetrack, featuresort}. Distinct SORT-based methods primarily differ in their hand-crafted rules for association across three key components:
(i) \textit{Tracklet Representation}: common approaches include mean~\cite{tracktor} or exponential moving averages of detection features \cite{TRMOT, fairmot}, or minimal distance to a feature bank~\cite{deepsort}. GHOST \cite{ghost} provides a comprehensive analysis of various "proxies" for computing the distance between a tracklet and a single detection, including the "Exponential Moving Average Feature Vector", "Median Feature Vector", "Last Frame Feature Vector", among others.
(ii) \textit{Feature Fusion}: methods range from weighted averaging of motion and appearance cues \cite{ghost} or additional cues \cite{hybridsort, featuresort}, to adaptive weighting schemes \cite{deepocsort} and threshold-based gating \cite{strongsort, botsort}. GHOST \cite{ghost} also conducts an extensive study examining how different "Motion Weight" values (weighting factors combining motion and appearance cost matrices) impact tracking performance across various datasets.
(iii) \textit{Multi-stage Matching}: trackers employ either single-stage \cite{botsort} or cascaded matching \cite{deepsort}, filtering candidates based on confidence scores \cite{bytetrack} or track age \cite{deepsort}, while using different cue at each stage. As described in \cref{sec:introduction}, Multi-stage matching involves computing distinct association cost matrices at each stage, using carefully selected subsets of active tracklets and detections (filtered by detection confidence or tracklet age). Each stage employs the Hungarian algorithm for bipartite matching, with unmatched tracklets/detections being processed in subsequent stages.

Most recent state-of-the-art methods typically implement a two-stage approach: an initial matching stage using custom heuristics (often incorporating ReID features), followed by a motion-based stage using IoU between Kalman Filter predicted bounding boxes and current detections, following SORT's \cite{sort} original design.
For example, DeepSORT performs multiple cascade matching stages using ReID features, processing tracklets in order of age, before concluding with SORT's standard Kalman Filter association stage.

Our method take a different direction and replaces these heuristics for data association with a unified trainable architecture, that better leverages all available tracking cues to produce context-aware disentangled representations to be matched in a single stage. 

\textit{Tracklet Life Cycle Management} represents another important family of heuristics in SORT-based pipelines, handling tracklet initialization, termination, and false positive detection filtering. While our work focuses on replacing association heuristics with a learned module, we maintain standard Life Cycle Management heuristics. Future extensions of CAMEL could potentially incorporate life cycle management through specialized state tokens representing tracklets to be paused, detections that should initiate new tracklets, and detections to be filtered as false positives. This capability represents a promising direction for future research.

\mysection{Tracking-by-Detection with Learned Association}

While some previous works have explored data-driven tracking through graph networks \cite{lns} or transformers \cite{gtr, sushi}, most operate offline, with only a few pioneering works attempting to integrate learned components into online TbD pipelines~\cite{transmot, tadn, busca, strn}. 
Our proposed CAMELTrack falls within this category of MOT methods.

Notably, TransMOT \cite{transmot} introduces a spatial-temporal encoder for tracklet representation and a transformer for feature fusion, but relies on a hand-crafted multi-stage matching pipeline where the learned components are only used in the second stage, while the first and third stages remain purely based on IoU and re-identification (ReID) heuristics.

TADN \cite{tadn} introduced a transformer-based decision network for learning tracklet-detection association with limited performance on MOTChallenge, likely related to their recursive training setup that cannot model hard association scenarios and accomodate for data augmentation like our association-centric training do. 
While BUSCA \cite{busca} proposes a decision transformer for associating tracklets with candidate detections, it serves only as a plug-in module for detection recovery on top of traditional TbD pipelines. 

STRN \cite{strn} introduced a Spatial-Temporal Relation Networks for data driven feature fusion, but their architectural design lacks the modularity to account for any type of input cue, and their pipeline still maintains other heuristic components.
While these works represent initial steps toward learned association, they still remain dependent on heuristics. In contrast, our approach makes a decisive break from hand-crafted rules by introducing a completely trainable association module

\mysection{Online Detection-by-Tracking} 

Recently, end-to-end (E2E) methods \cite{motip, motr, motrv2, trackformer, transtrack, memotr, comot, adatrack, tldmot, utm} following the Detection-by-Tracking (DbT) paradigm \cite{tracktor} have emerged as a promising, heuristic-free alternative to TbD approaches. Building upon DETR~\cite{detr} architecture, these methods jointly learn object detection and association, using track queries to re-detect past objects across frames. Despite their elegant design that learns association in a data-driven way similar to our approach, E2E still struggle to reach SotA performance on a wide range of datasets. This is because E2E methods face several limitations: (i) their detector-centric multi-frame training with short time windows struggles with long-term associations \cite{memot}, (ii) they lack TbD's modular ability to leverage specialized external models (e.g., ReID, motion, ...) \cite{motip},  (iii) the inherent conflict between detection and association objectives \cite{motrv2} in a shared model limits their overall performance and (iv) they require extensive training data and computational resources to achieve competitive performance (typically a few days on 8 GPUs \cite{motr}).
In contrast, our method focuses solely on learning an association strategy, requiring an order of magnitude less training compute, and maintains TbD's ability to leverage off-the-shelf detection, motion, and ReID models.

\begin{figure*}[p]
    \centering
     \begin{subfigure}[b]{0.48\linewidth}
         \centering
        \includegraphics[width=\textwidth]{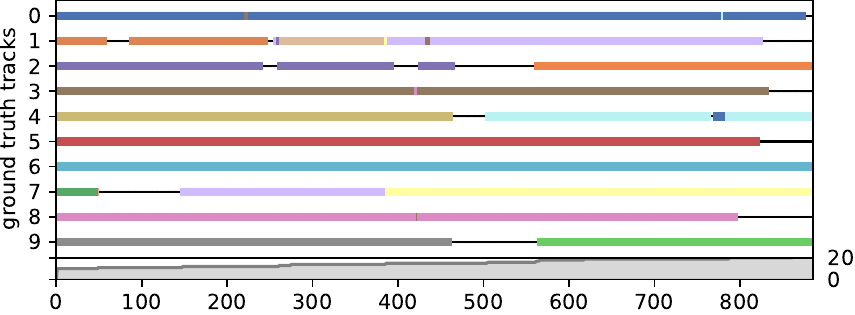}
         \caption{DiffMOT \cite{diffmot}}
     \end{subfigure}
     \hfill
     \begin{subfigure}[b]{0.48\linewidth}
         \centering
        \includegraphics[width=\textwidth]{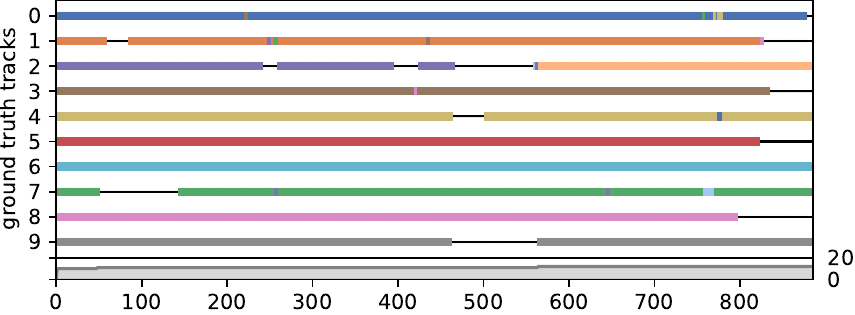}
         \caption{CAMEL} 
     \end{subfigure}
     \begin{subfigure}[b]{0.9\linewidth}
         \centering
            \includegraphics[width=\textwidth]{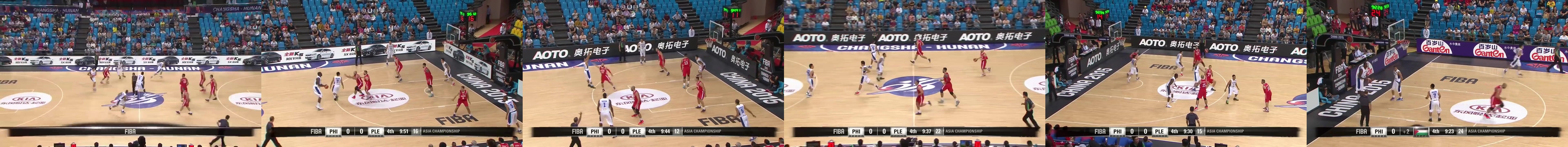}
         \caption{\textit{v\_4r8QL\_wglzQ\_c001} timeline}
     \end{subfigure}
     \begin{subfigure}[b]{0.48\linewidth}
         \centering
        \includegraphics[width=\textwidth]{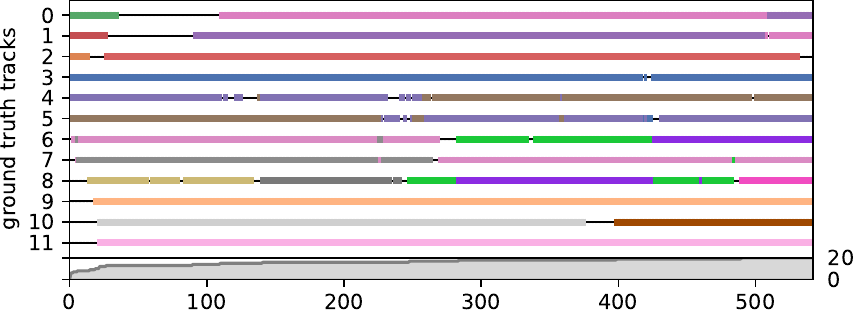}
         \caption{DiffMOT \cite{diffmot}}
     \end{subfigure}
     \hfill
     \begin{subfigure}[b]{0.48\linewidth}
         \centering
        \includegraphics[width=\textwidth]{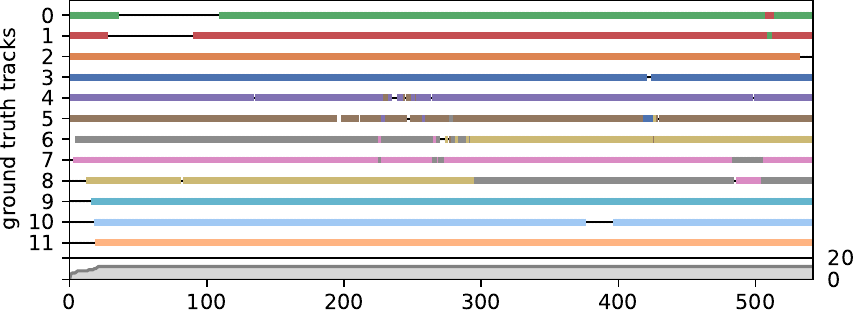}
         \caption{CAMEL}
     \end{subfigure}
    \begin{subfigure}[b]{0.9\linewidth}
         \centering
            \includegraphics[width=\textwidth]{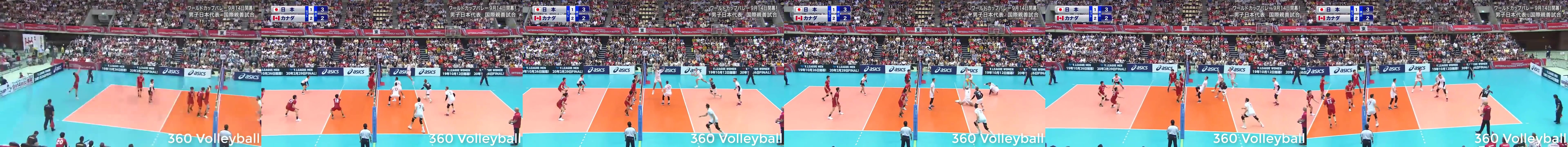}
         \caption{\textit{v\_0kUtTtmLaJA\_c004} timeline}
     \end{subfigure}
     \begin{subfigure}[b]{0.48\linewidth}
         \centering
        \includegraphics[width=\textwidth]{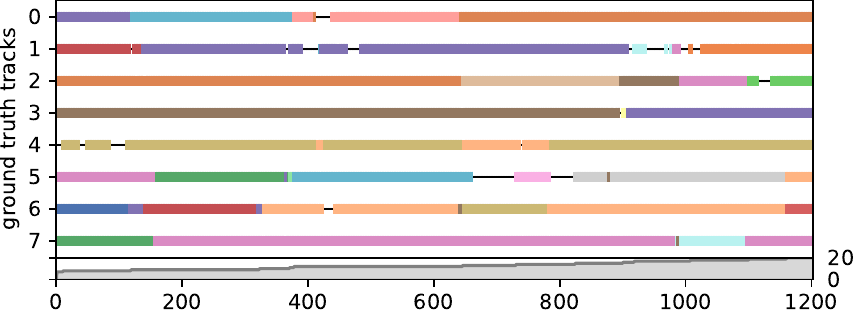}
         \caption{DiffMOT \cite{diffmot}}
     \end{subfigure}
     \hfill
     \begin{subfigure}[b]{0.48\linewidth}
         \centering
        \includegraphics[width=\textwidth]{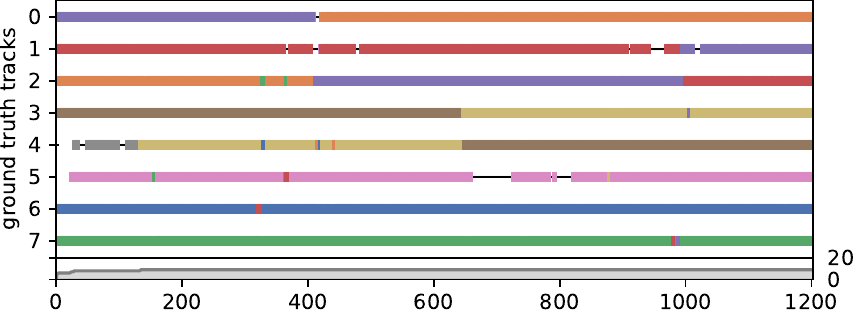}
         \caption{CAMEL}
     \end{subfigure}
    \begin{subfigure}[b]{0.9\linewidth}
         \centering
            \includegraphics[width=\textwidth]{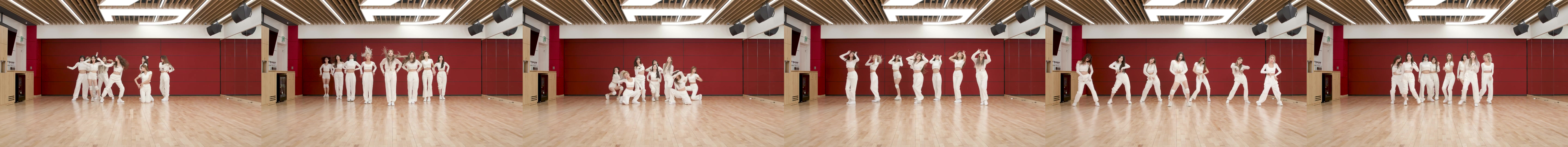}
         \caption{\textit{dancetrack0007} timeline}
     \end{subfigure}
     \begin{subfigure}[b]{0.48\linewidth}
         \centering
        \includegraphics[width=\textwidth]{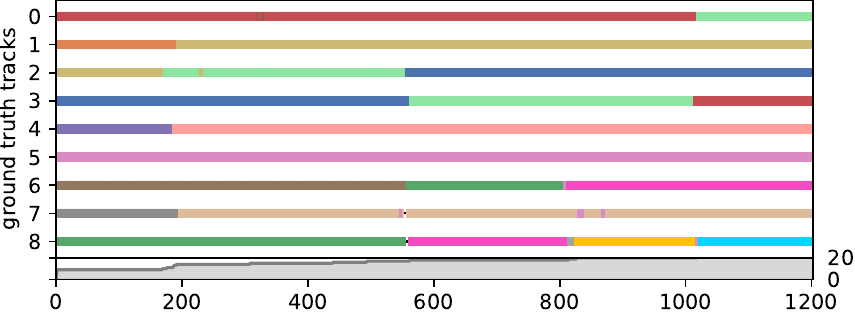}
         \caption{DiffMOT \cite{diffmot}}
     \end{subfigure}
     \hfill
     \begin{subfigure}[b]{0.48\linewidth}
         \centering
        \includegraphics[width=\textwidth]{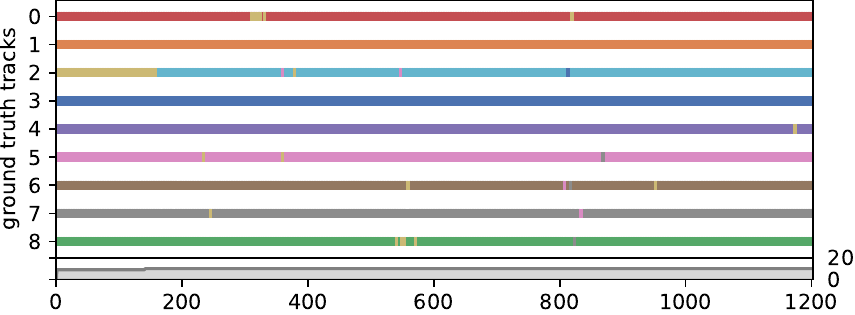}
         \caption{CAMEL}
     \end{subfigure}
    \begin{subfigure}[b]{0.9\linewidth}
         \centering
            \includegraphics[width=\textwidth]{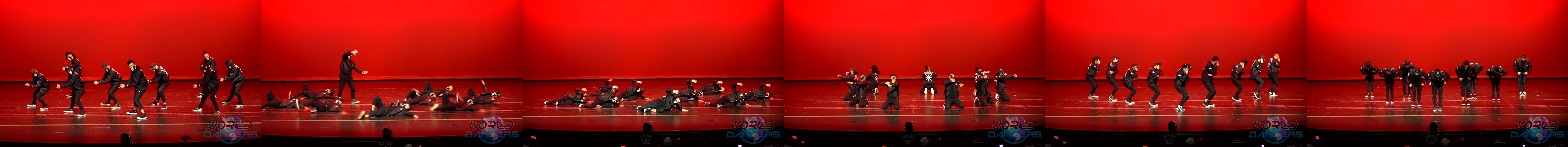}
         \caption{\textit{dancetrack0077} timeline}
     \end{subfigure}
    \caption{Visualization of tracking results on additional videos from the SportsMOT and DanceTrack validation sets. (a-c) video \textit{v\_4r8QL\_wglzQ\_c001} from SportsMOT. (d-f) video \textit{v\_0kUtTtmLaJA\_c004} from SportsMOT. (g-i) video \textit{dancetrack0007}. (j-l) video \textit{dancetrack0077}.  
    }
    \label{fig:qual-supp}
\end{figure*}
\section{Additional Qualitative Results} \label{sec:additional_qualitative_results}
\cref{fig:qual-supp} shows additional qualitative comparisons between CAMELTrack and DiffMOT in a timeline view. Using identical detections, we compare against DiffMOT which achieves near state-of-the-art performance on both DanceTrack and SportsMOT. These sequences, like those in \cref{fig:qualitative-results}, illustrate tracking behavior during challenging scenarios such as scene re-entries and occlusions.

\end{document}